\documentclass[10pt]{article} 
\usepackage[preprint]{tmlr}

\usepackage{amsmath,amsfonts,bm}

\def\eqref#1{equation~\ref{#1}}

\def\1{\bm{1}}

\DeclareMathAlphabet{\mathsfit}{\encodingdefault}{\sfdefault}{m}{sl}
\SetMathAlphabet{\mathsfit}{bold}{\encodingdefault}{\sfdefault}{bx}{n}

\usepackage{hyperref}
\usepackage{url}

\usepackage[T1]{fontenc}
\usepackage{graphicx}
\usepackage{booktabs}

\usepackage{microtype}
\usepackage{amsmath}
\usepackage{amssymb}
\usepackage{amsfonts}
\usepackage{dsfont}
\usepackage{algorithm}
\usepackage{algpseudocode}
\usepackage{xcolor}
\usepackage{colortbl}
\usepackage{multirow}
\usepackage{appendix} 
\usepackage{lscape} 

\usepackage{amsthm}
\newtheorem{definition}{Definition}

\title{DiffStyleTS: Diffusion Model for Style Transfer in Time Series}

\author{Mayank Nagda$^{1}$, Phil Ostheimer$^{1}$, Justus Arweiler$^{1}$, Indra Jungjohann$^{1}$, 
Jennifer Werner$^{2}$, Dennis Wagner$^{1}$, Aparna Muraleedharan$^{3}$, Pouya Jafari$^{1}$, 
Jochen Schmid$^{2}$, Fabian Jirasek$^{1}$, Jakob Burger$^{3}$, Michael Bortz$^{2}$, 
Hans Hasse$^{1}$, Stephan Mandt$^{4}$, Marius Kloft$^{1}$, Sophie Fellenz$^{1}$\\[0.75em]
$^{1}$RPTU Kaiserslautern-Landau\quad$^{2}$Fraunhofer ITWM\quad$^{3}$TU Munich\quad$^{4}$University of California
}

\def\month{MM}  
\def\year{YYYY} 
\def\openreview{\url{https://openreview.net/forum?id=XXXX}} 

\begin{document}

\maketitle

\begin{abstract}
Style transfer combines the content of one signal with the style of another. It supports applications such as data augmentation and scenario simulation, helping machine learning models generalize in data-scarce domains. While well developed in vision and language, style transfer methods for time series data remain limited. We introduce DiffTSST, a diffusion-based framework that disentangles a time series into content and style representations via convolutional encoders and recombines them through a self-supervised attention-based diffusion process. At inference, encoders extract content and style from two distinct series, enabling conditional generation of novel samples to achieve style transfer. We demonstrate both qualitatively and quantitatively that DiffTSST achieves effective style transfer. We further validate its real-world utility by showing that data augmentation with DiffTSST improves anomaly detection in data-scarce regimes.
\end{abstract}

\section{Introduction}

Despite their central role in monitoring and decision-making systems, time series data often remain limited precisely where they matter most—in safety-critical, high-cost, or privacy-sensitive domains. Accurate models of patient vitals, energy consumption, or market fluctuations, for example, require large and diverse datasets that are rarely available in practice. Machine learning models trained under such data scarcity frequently fail to generalize, hindering reliable forecasting and anomaly detection. This persistent challenge has fueled interest in generative approaches that can synthesize realistic yet diverse time series to enrich training data and support scenario simulation \citep{DBLP:conf/ijcai/Wen0YSGWX21}.

Among generative strategies, \emph{time series style transfer} (TSST) \citep{lahamtsst} offers a promising but underexplored direction. Inspired by its success in \emph{image} \citep{gatys2016image} and \emph{text} \citep{shen2017style} domains, TSST aims to synthesize new sequences by combining the \emph{content} of one series with the \emph{style} of another. In time series, \emph{content} typically captures the low-frequency global structure—such as daily demand cycles in energy data—while \emph{style} represents high-frequency local variations, like short-term fluctuations driven by consumer behavior. By separating and recombining these factors, TSST can generate realistic yet novel time series, offering a powerful tool for both \emph{simulation} and \emph{data augmentation}.

However, realizing TSST in practice remains challenging. Early attempts rely on simple signal manipulations such as overlaying or filtering, which often introduce temporal misalignments \citep{tebaldi2022stitches}. Neural approaches adapted from vision depend on handcrafted features or require paired data \citep{takemoto2019enriching,el2022styletime}, limiting their generality. Moreover, most existing methods are \emph{deterministic}, producing a single outcome rather than capturing the diversity of plausible futures—an essential property for generative modeling. These limitations call for a framework that can (i) disentangle content and style, (ii) recombine them in a principled way, and (iii) generate diverse, realistic sequences without relying on parallel data.

Recent advances in \emph{diffusion models} offer precisely these capabilities. By iteratively denoising random noise, diffusion models learn to approximate complex data distributions \citep{ho2020denoising}. They have revolutionized \emph{image} \citep{dhariwal2021diffusion}, \emph{video} \citep{ho2022imagen,singer2022makeavideo}, and \emph{audio} \citep{kong2021diffwave} generation, and have recently shown promise for \emph{image style transfer} \citep{wang2023stylediffusion}. We argue that their stochastic sampling, conditioning flexibility, and self-supervised training make diffusion models an ideal foundation for time series style transfer.

\begin{figure}
    \centering
    \includegraphics[scale=0.7]{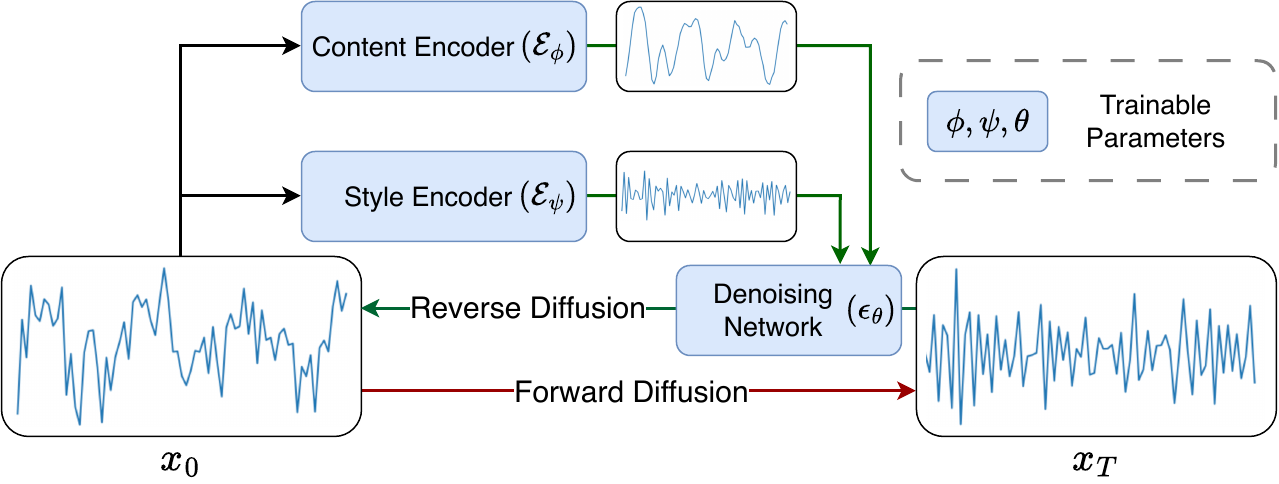}
    \caption{Overview of DiffTSST. The input series $x_{0}$ is gradually corrupted into $x_{T}$ through an iterative forward diffusion process. Content and style features are extracted by dedicated encoders $\mathcal{E}_{\phi}$ and $\mathcal{E}_{\psi}$. In the reverse diffusion process, a denoising network $\epsilon_{\theta}$ conditioned on these representations progressively reconstructs the input signal. At inference, content and style are drawn from distinct series, enabling the synthesis of new sequences via style transfer.}
    \label{fig:main-fig}
\end{figure}

We introduce \textbf{DiffTSST}, a novel \emph{diffusion-based framework for time series style transfer} that unifies disentanglement and generation within a single probabilistic model. Unlike existing deterministic or feature-engineered approaches, DiffTSST learns to separate and recombine the latent factors of \emph{content} and \emph{style} through a self-supervised diffusion process, eliminating the need for paired or annotated data. As shown in Figure~\ref{fig:main-fig}, DiffTSST consists of three tightly integrated components: (1) a forward diffusion process that progressively perturbs a time series with noise, (2) convolutional encoders that extract complementary content and style representations, and (3) a conditional denoising network that fuses these representations to reconstruct or synthesize new sequences. During training, DiffTSST uses each series as its own content–style pair, learning the underlying generative mechanism of temporal structure. At inference, content and style can be drawn from distinct series, enabling controllable and diverse style transfer that captures both global patterns and fine-grained dynamics.

Through extensive experiments across domains, we show that DiffTSST achieves realistic and diverse generations, and that its augmented data can significantly enhance downstream tasks such as anomaly detection in data-scarce regimes.

\textbf{Our contributions are threefold:}

\begin{itemize}
\item We formalize the problem of time series style transfer by establishing its core principles to guide future research.
\item We propose DiffTSST, the first diffusion-based approach to TSST, capable of generating diverse and realistic sequences beyond the limitations of deterministic methods.
\item Through extensive experiments across multiple domains, we demonstrate both qualitative fidelity and quantitative improvements, including gains in anomaly detection under data-scarce regimes.
\end{itemize}

The remainder of the paper is structured as follows: Section~\ref{sec:related} reviews related work, Section~\ref{sec:problem_setting} outlines the problem setting and background, Section~\ref{sec:methodology} details our approach, Section~\ref{sec:experiments} presents experiments, and Section~\ref{sec:conclusion} concludes the paper.

\section{Related Work}
\label{sec:related}

We review prior work on style transfer and diffusion models, highlighting the gaps our method addresses.

\subsection{Style Transfer}

Style transfer has been widely studied in computer vision and natural language processing. In vision, pioneering work demonstrated how content of an image can be recombined with style to synthesize novel images \citep{gatys2016image,luan2017deep}, while in text, stylistic attributes such as sentiment or formality can be modified without altering semantic meaning \citep{shen2017style,fu2018style}. Despite its success in these domains, style transfer for time series remains largely underexplored. Unlike images or text, time series data evolve continuously over time \citep{kantz2003nonlinear}, posing unique challenges. In time series, style refers to local variations, such as noise and volatility patterns, while content corresponds to global structure, such as long-term trends \citep{lahamtsst,yingzhen2018disentangled}. 

Existing TSST methods can be broadly grouped into heuristic, vision-based, and neural approaches. Heuristic methods simulate style transfer by relying on rule-based techniques such as overlaying high-frequency residuals onto smoothed trends \citep{tebaldi2022stitches,xu2024style}. While simple and computationally efficient, these methods often struggle with temporal misalignment, amplitude mismatches, and poor generalization to real-world time series data \citep{lahamtsst}. Vision-based approaches convert time series into visual forms such as recurrence plots or spectrograms, and then apply image-style transfer techniques \citep{takemoto2019enriching}. However, this representation is lossy, discards temporal ordering and reduces interpretability. Neural approaches attempt to learn content and style representations directly from time series \citep{el2022styletime}, but remain limited by reliance on handcrafted features or small paired datasets, constraining their ability to generalize. Overall, current methods for TSST are either heuristic, lossy, or deterministic, and do not provide a principled generative approach for diverse and realistic style transfer.

\subsection{Diffusion Models for Time Series} 

Diffusion models have recently emerged as state-of-the-art generative frameworks. The core idea of diffusion models is to gradually transform a simple noise distribution into a complex data distribution through a learned denoising process. Denoising Diffusion Probabilistic Models (DDPMs) and Denoising Diffusion Implicit Models (DDIMs) \citep{song2020score} have demonstrated state-of-the-art performance in high-fidelity image synthesis, outperforming established adversarial methods in terms of output quality and diversity. Beyond vision, diffusion models have also been explored in audio \citep{kong2020diffwave} and text \citep{gong2022diffuseq} generation tasks.

The application of diffusion models to time series is a relatively new frontier, with recent work exploring their use in imputation \citep{tashiro2021csdi}, forecasting \citep{rasul2021autoregressive}, controlled editing \citep{jing2024towards}, and extreme value modeling \citep{galib2024fide}. CSDI \citep{tashiro2021csdi} adapts score-based diffusion for time series imputation, leveraging conditional training to exploit observed correlations. TimeGrad \citep{rasul2021autoregressive} employs an autoregressive denoising diffusion approach for multivariate probabilistic forecasting, outperforming prior methods. TEdit \citep{jing2024towards} introduces a diffusion-based time series editing framework that modifies specified attributes while preserving other properties. FIDE \citep{galib2024fide} enhances generative modeling for time series by incorporating frequency-domain inflation and extreme value distributions to better capture rare events. 

Despite the success of diffusion models in time series generation \citep{tashiro2021csdi,jing2024towards}, their application to style transfer remains unexplored. Unlike vision and text, where pre-trained models and large-scale datasets facilitate style transfer, time series lacks such resources, making it a more challenging domain. To address this, we propose a diffusion-based TSST framework that decomposes time series into content and style and employs a Transformer-based \citep{vaswani2017attention} conditional denoising network to recombine them. This approach enables diverse and realistic style transfer while mitigating common challenges such as temporal misalignment and amplitude mismatch.

\section{Problem Setting and Background}
\label{sec:problem_setting}
In this section, we first formalize the task of time series style transfer by defining content and style decomposition, outlining key requirements, and presenting the problem formulation. We then provide background on diffusion models.

\subsection{Problem Setting}

An observed signal can generally be decomposed into two components: a content component that captures the global structure and a style component that introduces local variations. In the context of time series data, these components naturally emerge at different frequency scales: content is expressed in low-frequency global structure with long-range dependencies, while style manifests as high-frequency local variations with short-range dependencies. Here, dependencies refer to how strongly present values relate to past values across different time lags: long-range dependencies capture correlations that persist over extended horizons, whereas short-range dependencies reflect relationships confined to more immediate intervals. The precise scale of these dependencies can vary across domains, for instance, long-range in finance may span months, while in neuroscience it may cover only seconds.

In TSST, the aim is to generate a time series that preserves the content of one series while adopting the style of another. Building on style transfer research in NLP \citep{ostheimer-etal-2023-text}, we pose three key requirements for TSST: content preservation, style integration, and realism, as formally defined below.

\begin{definition}[Time Series Style Transfer (TSST)] Given a content time series $a \in \mathbb{R}^{L}$ and a style time series $b \in \mathbb{R}^{L}$, the objective of TSST is to learn a function $f: \mathbb{R}^{L} \times \mathbb{R}^{L} \to \mathbb{R}^{L}$ such that the generated time series $\hat{x} = f(a, b)$ satisfies the following requirements: 
\begin{itemize}
    \item (R1) Content Preservation: \( a \) and \( \hat{x} \) should share long-range dependencies and global structure.
    \item (R2) Style Integration: \( \hat{x} \) should exhibit local variations, noise characteristics, and volatility patterns resembling \( b \).
    \item (R3) Realism: \( \hat{x} \) should be an in-distribution sample with respect to the target domain, ensuring statistical plausibility.      
\end{itemize}
\end{definition}

Simultaneously integrating style without distorting the content’s long-range dependencies while ensuring realism is challenging, particularly in the absence of paired content--style data. The problem requires decomposing a time series into separate content and style components and recombining them within a generative model. Diffusion models are well suited to this setting, as they support conditioning on both components and enable diverse, realistic generation.

\subsection{Diffusion Models}
Diffusion models have emerged as powerful generative models by learning a structured reverse process to map Gaussian noise to complex data distributions. The diffusion process mainly consists of a forward noising step and a learned reverse denoising step.

\paragraph{Forward Process.} 
A time series \( x \in \mathbb{R}^{L} \) undergoes a Markovian Gaussian process, where noise is incrementally added over \( T \) steps, yielding a sequence \( x_0 \to x_1 \to \dots \to x_T \), where $x_0 = x$ is the original series and $x_t$ denotes the noisy series at step $t$. Given \( x_0 \), the distribution at any timestep \( t \) remains Gaussian:
\begin{equation*}
    q(x_t \mid x_0) = \mathcal{N} \big( x_t \mid \sqrt{\overline{\alpha}_t} x_0, (1 - \overline{\alpha}_t) I \big),
\end{equation*}
where \( \overline{\alpha}_t = \prod_{s=1}^{t} (1 - \beta_s) \) determines the noise schedule. As \( t \to T \), the data converges to an isotropic Gaussian distribution.

\paragraph{Reverse Process.}
To recover \( x_0 \), the model learns to denoise the sequence \( x_T \to x_{T-1} \to \dots \to x_0 \). While the true reverse transition \( q(x_{t-1} \mid x_t, x_0) \) is Gaussian with a known mean and variance, it requires access to \( x_0 \), making it intractable during inference. Instead, it is approximated by a learned model:
\begin{equation*}
    p_\theta(x_{t-1} \mid x_t) = \mathcal{N} \big( x_{t-1} \mid \mu_\theta(x_t, t), \sigma_t^2 I \big),
\end{equation*}
where \( \mu_\theta(x_t, t) \) is predicted by a neural network, and \( \sigma_t^2 \) is a predefined or learned variance. Starting from Gaussian noise \( \hat{x}_T \sim \mathcal{N}(0, I) \), the model iteratively refines the sample to obtain \( \hat{x}_0 \), an estimate of the original input \( x_0 \).

\paragraph{Training Objective.}
The denoising model \( \epsilon_\theta \) is trained to predict the added noise \( \epsilon \) using the objective:
\begin{equation*}
    \label{eq:loss}
    \mathcal{L} = \mathbb{E}_{x_0, \epsilon, t} \Big[ \big\| \epsilon - \epsilon_\theta(\sqrt{\overline{\alpha}_t} x_0 + \sqrt{1 - \overline{\alpha}_t} \epsilon, t) \big\|^2 \Big].
\end{equation*}

\paragraph{Sampling.}
Once trained, a diffusion model generates samples by starting from Gaussian noise 
\(x_T \sim \mathcal{N}(0, I)\) and iteratively applying the learned reverse transitions. 
At each step \(t\), the model predicts the injected noise 
\(\epsilon_\theta(x_t, t)\), which is used to estimate the clean signal and update the sample:
\begin{equation*}
    x_{t-1} = \frac{1}{\sqrt{\alpha_t}} 
    \left( x_t - \frac{1 - \alpha_t}{\sqrt{1 - \overline{\alpha}_t}} \, 
    \epsilon_\theta(x_t, t) \right) + \sigma_t z, 
    \quad z \sim \mathcal{N}(0, I).
\end{equation*}
Iterating this process from \(t=T\) down to \(t=1\) yields \(x_0\), 
a synthetic sample from the data distribution.

\paragraph{Conditional Diffusion.}
Diffusion models can be extended to conditional generation by incorporating auxiliary information into the denoising process. Instead of generating samples unconditionally, the model is guided by side information such as class labels, reference signals, or learned embeddings. The conditional reverse distribution is then expressed as:
\begin{equation*}
    p_\theta(x_{t-1} \mid x_t, c) = \mathcal{N}( x_{t-1} \;\big|\; \mu_\theta(x_t, t, c), \, \sigma_t^2 I ),
\end{equation*}
where the conditioning variable \( c \) shapes the reverse trajectory, steering generation toward outputs consistent with the desired attributes while retaining stochasticity.

\paragraph{Classifier-Free Guidance.}
A key challenge in conditional diffusion is balancing adherence to conditioning signals with sample diversity. Classifier-free guidance (CFG) \citep{ho2022classifier} addresses this by training the denoiser jointly on conditional and unconditional objectives. During training, conditioning inputs are randomly dropped with probability 
\(p_{\text{drop}}\), enabling the model to learn both conditional 
\(\epsilon_\theta(x_t, t, c)\) and unconditional \(\epsilon_\theta(x_t, t)\) 
denoising functions.
At inference, the two predictions are combined through a guidance scale \(s \geq 1\):
\begin{equation*}
    \epsilon_\theta(x_t, t, c) = \epsilon_\theta(x_t, t) 
    + s \cdot \Big( \epsilon_\theta(x_t, t, c) - \epsilon_\theta(x_t, t) \Big).
\end{equation*}
The modified prediction is then used in the sampling update, 
enforcing stronger adherence to conditioning when \(s > 1\) while maintaining diversity for smaller \(s\).

\begin{figure}[!t]
    \centering
    \includegraphics[scale=0.75]{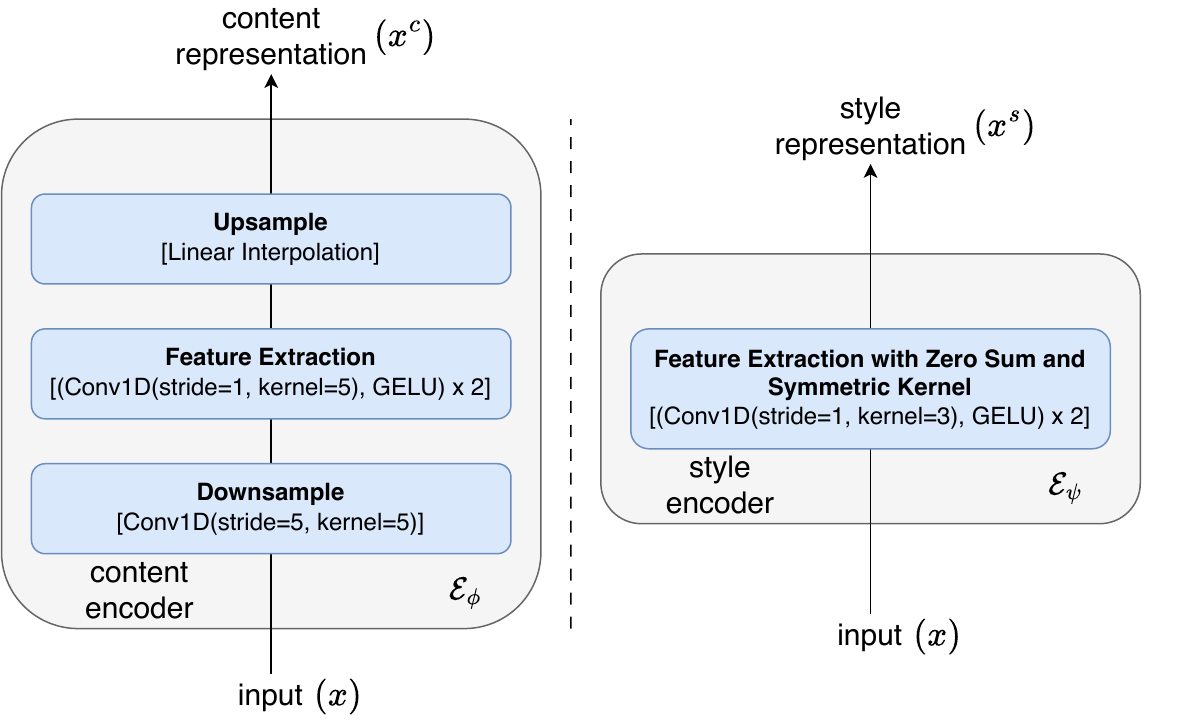}
    \caption{Content and style encoders.  
Left: The content encoder captures the low-frequency content by strongly downsampling the input, processing it at reduced resolution, and then upsampling to the original length, preserving global trends while discarding fine details.  
Right: The style encoder captures the high-frequency style using small convolutional filters constrained to avoid constant offsets and phase shifts, ensuring sensitivity to local fluctuations.  
Together, the two encoders disentangle global structure from local variations in the time series.}
    \label{fig:encoder_decoder}
\end{figure}

\section{Methodology}
\label{sec:methodology}
This section introduces the DiffTSST architecture along with its training and inference routines. We denote the reference style series as $a$ and the content series as $b$. During training, a single time series $x$ is used for both $(x = a = b)$, while during inference, distinct time series $a$ and $b$ are provided for style and content to enable style transfer.

\subsection{Extracting Style and Content from a Time Series}

We decompose a time series into content and style components based on their frequency characteristics and temporal dependencies. To achieve this, we design two complementary encoders (Figure~\ref{fig:encoder_decoder}): a content encoder that captures low-frequency global structure with long-range dependencies, and a style encoder that captures high-frequency local variations with short-range dependencies.

\paragraph{Content Encoder.}  
The content encoder $\mathcal{E}_{\phi}(\cdot)$ maps an input time series $x \in \mathbb{R}^L$ to a content representation $x^c \in \mathbb{R}^L$. It operates in three stages. First, a downsampling module applies a convolution with a sufficiently large kernel and stride, acting as a learnable low-pass filter that shortens the sequence and suppresses high-frequency style information. Next, a feature extraction stage with convolutional blocks and large kernels captures long-range dependencies efficiently at the reduced resolution. Finally, a projection–upsampling stage restores the original length through linear interpolation while projecting back to a single channel. This design isolates the low-frequency global structure while filtering out high-frequency variations.

\paragraph{Style Encoder.}  
The style encoder $\mathcal{E}_{\psi}(\cdot)$ maps an input time series $x \in \mathbb{R}^L$ to a style representation $x^s \in \mathbb{R}^L$. Unlike the content encoder, it operates directly at the original resolution without downsampling. It applies a stack of convolutional layers with small kernels, which are well suited for capturing short-range dependencies. To prevent the filters from reintroducing global trends or phase shifts, we constrain each kernel to have zero mean (removing constant offsets) and to be symmetric (enforcing linear-phase behavior). Reflection padding is used to preserve sequence length. This design enables the style encoder to function as a learnable high-pass filter that emphasizes local variations while suppressing global structure.

\subsection{Denoising Diffusion Transformer for Style Transfer}
\begin{figure}[!t]
    \centering
    \includegraphics[scale=0.75]{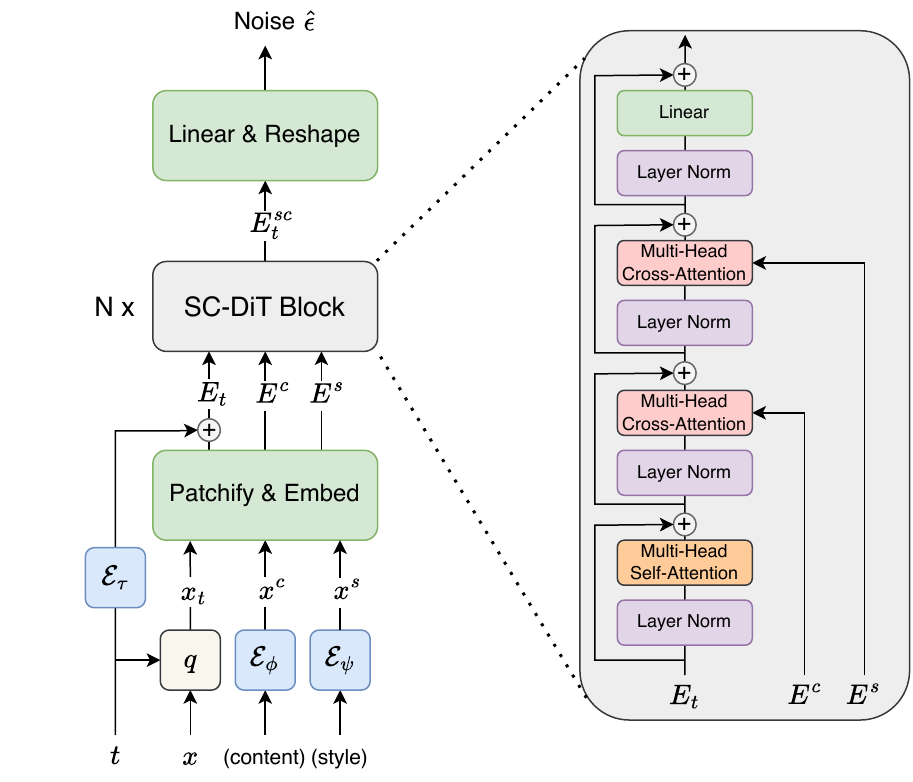}
    \caption{Architecture of the proposed diffusion transformer for time series style transfer. A noisy input $x_t$ is patchified and embedded, combined with a timestep embedding, and processed by a stack of SC-DiT blocks. Each block integrates information from both content and style encoders via cross-attention, while self-attention preserves temporal structure. The network predicts the residual noise $\hat{\epsilon}$ to guide reverse diffusion.}
    \label{fig:arch}
\end{figure}

Our denoising network, illustrated in Figure~\ref{fig:arch}, adopts a transformer-based architecture derived from diffusion transformers (DiTs) \citep{peebles2023scalable} for time series. During the forward diffusion process, Gaussian noise $\epsilon$ progressively corrupts the input series $x$, producing noisy sample $x_t = q(x, t) \in \mathbb{R}^L$ at step $t$. The denoising network learns to predict the added noise $\hat{\epsilon} \in \mathbb{R}^L$ directly from $x_t$. Cross-attention layers condition this prediction on content and style features, $x^c$ and $x^s$. We next describe each module in detail.

\paragraph{Patch Embedding.}
The noisy sample $x_t$, content $x^c$, and style $x^s$ in $\mathbb{R}^L$ are divided into $n=\frac{L}{p}$ non-overlapping patches of length $p$. A linear projection maps each patch into an embedding space of dimension $h$, producing patch embeddings $E_t, E^c, E^s \in \mathbb{R}^{n \times h}$, respectively.

\paragraph{Time Embedding.}
The diffusion step $t$ is encoded as a sinusoidal vector $\tau_t \in \mathbb{R}^h$, following the positional encoding scheme of \citet{vaswani2017attention}. We first define $\frac{h}{2}$ frequencies 
$\omega_j = \exp\!\big(-j \cdot \log(10000)/(\frac{h}{2}-1)\big)$ for $j=0,\dots,\frac{h}{2}-1$. The embedding is then constructed as $\tau_t = \big[\sin(t \cdot \omega_0), \cos(t \cdot \omega_0), \dots, \sin(t \cdot \omega_{\frac{h}{2}-1}), \cos(t \cdot \omega_{\frac{h}{2}-1})\big]$. $\tau_t$ is added to each patch in $E_t$ to provide explicit conditioning on the noise level.

\paragraph{SC-DiT Blocks.}
The core of the denoiser consists of stacked \emph{Style-Content Diffusion Transformer} (SC-DiT) blocks. Each block processes $E_t$, $E^c$, and $E^s$ through self-attention and cross-attention, following the standard transformer formulation \citep{vaswani2017attention}. For clarity, we describe a single attention head. Queries, keys, and values are defined as $Q = E W^Q$, $K = E W^K$, $V = E W^V$ for input $E$, with learnable $W^Q, W^K, W^V \in \mathbb{R}^{h \times h}$ and $d_k=h$. Self-attention models temporal dependencies within $E_t$ via
\[
\text{Attn}(E_t) = \text{Softmax}\!\left(\frac{QK^\top}{\sqrt{d_k}}\right)V.
\]
Cross-attention conditions the denoising on external signals: content information is incorporated from $E^c$ using $Q = E_t W^Q$, $K = E^c W^K$, $V = E^c W^V$, and style information from $E^s$ analogously. Each block also includes residual connections, normalization, and feed-forward layers. Stacking multiple SC-DiT blocks allows coherent fusion of content and style features. The output of this stage is denoted $E_t^{sc} \in \mathbb{R}^{n \times h}$.

\paragraph{Noise Prediction.}
The output embeddings $E_t^{sc} \in \mathbb{R}^{n \times h}$ are linearly projected back to the patch dimension $p$. The $n = \frac{L}{p}$ patches are then concatenated to reconstruct a sequence of length $L$, yielding the noise estimate $\hat{\epsilon} \in \mathbb{R}^L$.

\paragraph{Positional Encodings.}
We incorporate relative positional information using ALiBi (Attention with Linear Biases) \citep{press2022train}. Rather than adding positional vectors to the input patches, ALiBi modifies the attention logits with linear biases applied before the softmax. For positions $i$ and $j$, the bias is $\text{bias}(i,j) = -\alpha |i-j|$, where $\alpha$ is a predefined slope. In self-attention, this depends on pairwise patch distances; in cross-attention, on query–key distances. By encoding order directly in the attention scores, ALiBi avoids explicit embeddings and generalizes well to sequences longer than those seen during training, making it particularly suitable for TSST.

\subsection{Training and Inference}

\begin{algorithm}[t!]
\caption{Training with Classifier-Free Guidance}
\label{alg:training}
\begin{algorithmic}[1]
\State \textbf{Input:} dataset $\mathcal{D}$, steps $T$, schedule $\{\beta_t\}$, drop probabilities $p_c,p_s$
\State Initialize $\alpha_t=1-\beta_t$ and $\bar\alpha_t=\prod_{s=1}^t \alpha_s$
\Repeat
    \State Sample $x_0 \sim \mathcal{D}$, $t \sim \mathcal{U}\{1,\dots,T\}$, $\epsilon \sim \mathcal{N}(0,I)$
    \State Forward noising: $x_t=\sqrt{\bar\alpha_t}\,x_0+\sqrt{1-\bar\alpha_t}\,\epsilon$
    \State Compute $x^c = \mathcal{E}_\phi(x_0);\quad x^s = \mathcal{E}_\psi(x_0)$
    \State Apply random dropout: $\tilde{x}^c = x^c$ w.p.\ $1-p_c$, else $\varnothing$; \quad $\tilde{x}^s = x^s$ w.p.\ $1-p_s$, else $\varnothing$
    \State Take gradient step on: $\nabla_\theta \big\|\epsilon - \epsilon_\theta(x_t,t,\tilde{x}^c,\tilde{x}^s)\big\|^2$
\Until{convergence}
\end{algorithmic}
\end{algorithm}

\begin{algorithm}[t!]
\caption{Inference with Guidance Combination}
\label{alg:inference}
\begin{algorithmic}[1]
\State \textbf{Input:} content $a$, style $b$, steps $T$, schedule $\{\alpha_t\}$, guidance scales $s_c,s_s$, temperature $\lambda$
\State Initialize $x_T \sim \mathcal{N}(0,I)$
\State Compute $a^c = \mathcal{E}_\phi(a);\quad b^s = \mathcal{E}_\psi(b)$
\For{$t=T$ to $1$}
    \State Compute noise predictions: $\epsilon_u=\epsilon_\theta(x_t,t),\;\; \epsilon_c=\epsilon_\theta(x_t,t,a^c),\;\; \epsilon_s=\epsilon_\theta(x_t,t,b^s)$
    \State Combine: $\hat{\epsilon}=\epsilon_u+s_c(\epsilon_c-\epsilon_u)+s_s(\epsilon_s-\epsilon_u)$
    \State Update sample:
    \[
        x_{t-1}=\tfrac{1}{\sqrt{\alpha_t}}\!\left(\hat{x}_t-\tfrac{1-\alpha_t}{\sqrt{1-\bar{\alpha}_t}}\epsilon\right)+\lambda \sigma_t z, \quad z\!\sim\!\mathcal{N}(0,I) \text{ if }t>0, \text{ else } z=0
    \]
\EndFor
\State \textbf{Output:} $\hat{x} := x_0$
\end{algorithmic}
\end{algorithm}

The training procedure is outlined in Algorithm~\ref{alg:training}. For each sample $x_0 \sim \mathcal{D}$, the forward diffusion process produces a noisy input $x_t$. The content and style encoders, $\mathcal{E}_\phi$ and $\mathcal{E}_\psi$, then extract representations $x^c$ and $x^s$ from $x_0$. During training, both encoders operate on the same series, enabling the model to learn in a self-supervised manner without paired data. The denoiser $\epsilon_\theta$ is trained to recover the injected noise $\epsilon$ from $x_t$, conditioned on $(x^c, x^s)$. To allow flexible conditioning, the training employs classifier-free guidance: the content and style inputs are independently dropped with probabilities $(p_c, p_s)$.  

The inference procedure (Algorithm~\ref{alg:inference}) generates a new sequence by denoising a Gaussian sample $x_T \sim \mathcal{N}(0,I)$ over $T$ steps. Distinct inputs are used for content ($a$) and style ($b$), encoded as $a^c = \mathcal{E}_\phi(a)$ and $b^s = \mathcal{E}_\psi(b)$. At each step, the model predicts unconditional, content-conditioned, and style-conditioned noise estimates, which are combined using guidance scales $(s_c, s_s)$ to form $\hat{\epsilon}$. This estimate drives the reverse diffusion process, where the temperature parameter $\lambda$ scales the Gaussian noise injected at each step, controlling the diversity of generated samples. The process progressively refines $x_T$ into the final output $\hat{x}$, which preserves the content of $a$ while adopting the style of $b$.

\section{Experiments}
\label{sec:experiments}
In this section, we first describe the datasets, comparison models, and evaluation metrics used in our experiments. We then benchmark the proposed DiffTSST method against baselines and explore real-world use cases.

\subsection{Experimental Setup}
\paragraph{Datasets and Training.} The proposed DiffTSST model is trained using unpaired time series data, allowing it to generalize across diverse datasets. We train it on the MONASH time series corpus \citep{godahewa2021monash}, which comprises over 300K time series from about 30 different domains. Training is conducted in batches for 200K iterations, with 500 diffusion timesteps ($T$). Each batch consists of 256 randomly sampled time series of length 128. The model is trained on NVIDIA A100 GPUs, and both the model weights and training scripts are made available. The exact model architecture, training procedure, and hyperparameters are detailed in Appendix~\ref{app:arch}, Appendix~\ref{app:train}, and Appendix~\ref{app:hyper}, respectively.

\paragraph{Baselines.} 
We benchmark our method against three baselines: Wavelets, Stitching, and a neural style transfer (NST) adaptation from computer vision.  

\emph{Wavelets} \citep{yoo2019photorealistic,chaovalit2011discrete}: wavelet decompositions have been widely used in image style transfer  \citep{yoo2019photorealistic} and also time series analysis \citep{chaovalit2011discrete} to manipulate multi-resolution features. Here, we adopt the Haar transform to decompose both content and style series, swap the high-frequency components (treated as style), and apply the inverse transform to reconstruct the stylized signal.  

\emph{Stitching} \citep{tebaldi2022stitches,xu2024style}: 
the idea of stitching signal segments has been explored in climate simulations \citep{tebaldi2022stitches} and image synthesis \citep{xu2024style}. For time series, we smooth a sequence using an average filter. The residual (original minus smoothed signal) is treated as style. Style transfer is then performed by combining the smoothed content of one series with the residual of another to produce a blended sequence.  

\emph{Neural Style Transfer (NST)} \citep{gatys2016image}: we adapt the optimization-based framework from vision to time series. Here, the generated series $\hat{x}$ is treated as a learnable variable and optimized to simultaneously preserve content and match style statistics. Specifically, we minimize a weighted sum of distances in the embedding space:
\[
    \mathcal{L}(\hat{x}; a, b) 
    = \alpha \, \| f_c(\hat{x}) - f_c(a) \|^2 
    + \beta \, \| f_s(\hat{x}) - f_s(b) \|^2,
\]
where $f_c(\cdot)$ and $f_s(\cdot)$ denote content and style feature extractors, and $\alpha, \beta$ control the trade-off. This mirrors the neural style transfer setup in vision, but applied to the temporal domain. For feature extraction, we first decompose each series into content and style components using stitching, and then obtain their representations with the MOMENT foundation model \citep{goswami2024moment}.

\begin{figure}[t!]
    \centering
    \includegraphics[width=0.9\textwidth]{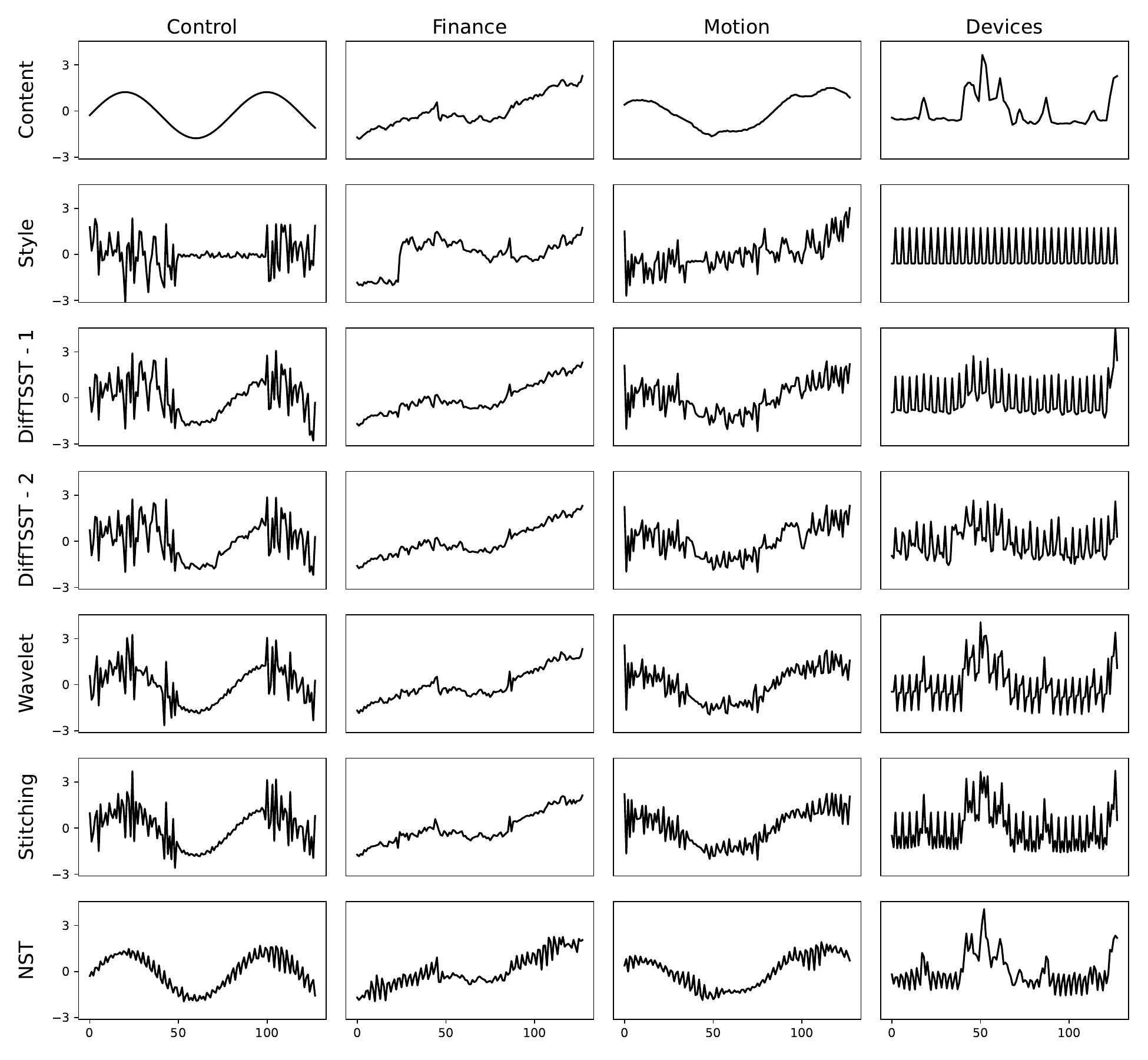}
    \caption{Qualitative comparison of style transfer methods across four content–style pairs (Control, Finance, Motion, Devices). Rows show the content, style, and outputs from DiffTSST (ours), wavelet, stitching, and neural optimization baselines. Only DiffTSST effectively integrates style dynamics into the generated series while preserving the underlying content, whereas baseline methods tend to either overfit to content or introduce artifacts.}
    \label{fig:qualitative}
\end{figure}

\subsection{Evaluation Metrics}

In the absence of paired content–style datasets, we evaluate style transfer by decomposing the generated series $\hat{x}$, the reference content series $a$, and the reference style series $b$. We apply a multi-scale averaging filter with kernel sizes $K = (3,5,15)$, which separates each signal into a smoothed content component and residual style components:
\begin{equation}
\begin{aligned}
    \hat{x}_{\text{smooth}}^{(i)} &= \operatorname{AvgFilter}_{K_i}\!\bigl( \hat{x}_{\text{smooth}}^{(i-1)} \bigr), \quad i=1,\dots,|K|, \\
    \hat{x}_{\text{res}}^{(i)}    &= \hat{x}_{\text{smooth}}^{(i-1)} - \hat{x}_{\text{smooth}}^{(i)},
\end{aligned}
\end{equation}
where $\operatorname{AvgFilter}_{L}(\hat{x}) = \hat{x} * \tfrac{1}{L}\mathds{1}_{L}$ denotes convolution with a uniform kernel of length $L$, $\hat{x}_{\text{smooth}}^{(0)}=\hat{x}$, the final smoothed signal $C(\hat{x})=\hat{x}_{\text{smooth}}^{(|K|)}$ represents the content, and the aggregated residuals $S(\hat{x})=\sum_{i=1}^{|K|} \hat{x}_{\text{res}}^{(i)}$ represent style.

We evaluate TSST using three complementary metrics: content preservation (CP), style integration (SI), and realism (RM), aligned with the requirements posed in Section~\ref{sec:problem_setting}. In all cases, we compute mean squared error (MSE) as the distance measure.

\paragraph{Content Preservation (CP).}
Measures how well the generated series $\hat{x}$ preserves the global structure of the content reference $a$:
\[
    \text{CP}(\hat{x},a) = \frac{1}{L} \sum_{t=1}^{L} (C(\hat{x})_t - C(a)_t)^2.
\]
Lower CP indicates stronger preservation of content.

\paragraph{Style Integration (SI).}
Measures how well the local variability of the style reference $b$ is captured:
\[
    \text{SI}(\hat{x},b) = \frac{1}{L} \sum_{t=1}^{L} (S(\hat{x})_t - S(b)_t)^2.
\]
Lower SI indicates closer alignment with the target style.

\paragraph{Realism (RM).}
Assesses whether the generated series resembles realistic data by comparing embeddings extracted with the MOMENT foundation model \citep{goswami2024moment}:
\[
    \text{RM}(\hat{x},a,b) = \tfrac{1}{2}\Bigl[\; \text{MSE}\bigl(f(C(\hat{x})), f(C(a))\bigr) + \text{MSE}\bigl(f(S(\hat{x})), f(S(b))\bigr) \;\Bigr],
\]
where $f(\cdot)$ is the embedding function. Lower RM indicates that generated samples lie closer to the real data manifold. A detailed discussion of evaluation metrics is provided in Appendix~\ref{app:eval}.

\begin{table*}[t]
\centering
\caption{Quantitative evaluation of time series style transfer methods. DiffTSST outperforms existing baselines on 19/23 datasets. Reported are overall scores (mean of CP, SI, RM per sample). Lower is better.}
\label{tab:overall}
\begin{tabular}{l|cccc}
\hline
Dataset & Stitching & Wavelet & NST & DiffTSST \\
\hline
Electricity  Demand & $0.07 \, \pm \, 0.01$ & $0.08 \, \pm \, 0.01$ & $0.12 \, \pm \, 0.01$ & $\mathbf{0.07} \, \pm \, \mathbf{0.01}$ \\
Bitcoin & $0.06 \, \pm \, 0.00$ & $0.07 \, \pm \, 0.00$ & $0.11 \, \pm \, 0.00$ & $\mathbf{0.05} \, \pm \, \mathbf{0.00}$ \\
Covid Deaths & $0.08 \, \pm \, 0.00$ & $0.09 \, \pm \, 0.01$ & $0.13 \, \pm \, 0.01$ & $\mathbf{0.07} \, \pm \, \mathbf{0.00}$ \\
FRED-MD & $0.04 \, \pm \, 0.00$ & $0.05 \, \pm \, 0.00$ & $0.09 \, \pm \, 0.00$ & $\mathbf{0.04} \, \pm \, \mathbf{0.00}$ \\
Kaggle Web Traffic & $0.08 \, \pm \, 0.00$ & $0.09 \, \pm \, 0.00$ & $0.13 \, \pm \, 0.00$ & $\mathbf{0.05} \, \pm \, \mathbf{0.00}$ \\
KDD Cup 2018 & $0.11 \, \pm \, 0.00$ & $0.13 \, \pm \, 0.00$ & $0.20 \, \pm \, 0.01$ & $\mathbf{0.08} \, \pm \, \mathbf{0.00}$ \\
London Smart Meters & $\mathbf{0.16} \, \pm \, \mathbf{0.00}$ & $0.17 \, \pm \, 0.00$ & $0.29 \, \pm \, 0.00$ & $0.19 \, \pm \, 0.00$ \\
NN5 Daily & $0.19 \, \pm \, 0.01$ & $0.26 \, \pm \, 0.01$ & $0.45 \, \pm \, 0.01$ & $\mathbf{0.12} \, \pm \, \mathbf{0.00}$ \\
Oikolab Weather & $0.14 \, \pm \, 0.01$ & $0.19 \, \pm \, 0.01$ & $0.24 \, \pm \, 0.01$ & $\mathbf{0.12} \, \pm \, \mathbf{0.01}$ \\
Pedestrian Counts & $0.16 \, \pm \, 0.00$ & $0.19 \, \pm \, 0.00$ & $0.25 \, \pm \, 0.00$ & $\mathbf{0.10} \, \pm \, \mathbf{0.00}$ \\
Rideshare & $0.16 \, \pm \, 0.01$ & $0.18 \, \pm \, 0.01$ & $0.25 \, \pm \, 0.02$ & $\mathbf{0.14} \, \pm \, \mathbf{0.01}$ \\
Saugeenday & $0.09 \, \pm \, 0.00$ & $0.10 \, \pm \, 0.00$ & $0.15 \, \pm \, 0.00$ & $\mathbf{0.06} \, \pm \, \mathbf{0.00}$ \\
Solar (10min) & $0.07 \, \pm \, 0.00$ & $0.08 \, \pm \, 0.01$ & $0.12 \, \pm \, 0.01$ & $\mathbf{0.07} \, \pm \, \mathbf{0.01}$ \\
Solar (4sec) & $\mathbf{0.05} \, \pm \, \mathbf{0.00}$ & $0.05 \, \pm \, 0.00$ & $0.09 \, \pm \, 0.00$ & $0.05 \, \pm \, 0.00$ \\
Sunspot & $0.18 \, \pm \, 0.01$ & $0.22 \, \pm \, 0.01$ & $0.30 \, \pm \, 0.02$ & $\mathbf{0.13} \, \pm \, \mathbf{0.01}$ \\
Temperature (rain) & $\mathbf{0.17} \, \pm \, \mathbf{0.00}$ & $0.26 \, \pm \, 0.00$ & $0.42 \, \pm \, 0.00$ & $0.17 \, \pm \, 0.01$ \\
Tourism (monthly) & $0.11 \, \pm \, 0.00$ & $0.14 \, \pm \, 0.00$ & $0.29 \, \pm \, 0.00$ & $\mathbf{0.10} \, \pm \, \mathbf{0.00}$ \\
Traffic (hourly) & $0.24 \, \pm \, 0.01$ & $0.28 \, \pm \, 0.01$ & $0.33 \, \pm \, 0.01$ & $\mathbf{0.20} \, \pm \, \mathbf{0.01}$ \\
US Births & $0.20 \, \pm \, 0.00$ & $0.28 \, \pm \, 0.00$ & $0.40 \, \pm \, 0.00$ & $\mathbf{0.16} \, \pm \, \mathbf{0.00}$ \\
Vehicle Trips & $0.19 \, \pm \, 0.00$ & $0.26 \, \pm \, 0.00$ & $0.38 \, \pm \, 0.00$ & $\mathbf{0.12} \, \pm \, \mathbf{0.00}$ \\
Weather & $0.12 \, \pm \, 0.00$ & $0.16 \, \pm \, 0.00$ & $0.23 \, \pm \, 0.00$ & $\mathbf{0.10} \, \pm \, \mathbf{0.00}$ \\
Wind (4sec) & $0.05 \, \pm \, 0.00$ & $0.06 \, \pm \, 0.00$ & $0.10 \, \pm \, 0.00$ & $\mathbf{0.05} \, \pm \, \mathbf{0.00}$ \\
Wind Farms (min) & $\mathbf{0.05} \, \pm \, \mathbf{0.00}$ & $0.09 \, \pm \, 0.00$ & $0.08 \, \pm \, 0.00$ & $0.06 \, \pm \, 0.00$ \\
\hline
Overall & $0.12 \, \pm \, 0.01$ & $0.15 \, \pm \, 0.01$ & $0.22 \, \pm \, 0.02$ & $\mathbf{0.10} \, \pm \, \mathbf{0.01}$ \\
\hline
\end{tabular}
\end{table*}

\subsection{Qualitative Evaluation}

A qualitative analysis is conducted on unseen styles to evaluate the efficacy of DiffTSST in zero-shot time series style transfer. The styles used in these experiments are not included in the training data and are sourced from the HIVE corpus \citep{middlehurst2021hive}, demonstrating the model’s ability to generalize to out-of-distribution styles.

Figure~\ref{fig:qualitative} compares outputs against baselines. Both wavelet and stitching methods largely superimpose superficial fluctuations from the style series onto the content, without adequately decomposing and integrating stylistic patterns. This results in outputs that remain content-dominant and struggle to capture distinctive style characteristics, for example, the abrupt jumps in the Control style series (around timestep 25-40) is ignored by all baselines but reproduced by DiffTSST. Similarly, in the Devices example, oscillatory spikes in the style lead wavelet and stitching to inject compensatory negative artifacts, rather than transferring the true stylistic structure. The NST baseline, adapted from vision tasks, preserves content but repeatedly inserts similar motifs, reflecting weak style sensitivity across all domains. In contrast, DiffTSST reproduces salient stylistic structures while maintaining global content trajectories. It also allows multiple diverse outputs per content–style pair as evidenced by multiple generatons (DiffTSST-1, and -2). Furthermore, ablations in Appendix~\ref{app:ablations_diversity} show that temperature parameter $\lambda$ controls diversity, enabling varied yet realistic outputs.

\subsection{Time Series Style Transfer Benchmark}

We evaluate style transfer performance using test splits from the Monash Time Series Forecasting Archive \citep{godahewa2021monash} as content references. For style references, we sample from five datasets in the HIVE corpus \citep{middlehurst2021hive}, which are excluded from training to ensure unseen styles at evaluation.  

Table~\ref{tab:overall} summarizes overall scores, computed as the mean of CP, SI, and RM per sample (lower is better). DiffTSST achieves the best performance on 19 out of 23 Monash datasets, showing strong generalization across diverse domains. A closer breakdown in Appendix~\ref{app:quantitative} reveals that deterministic baselines (Stitching, Wavelets) often achieve near-zero CP (Table~\ref{tab:content_preservation}), since they directly reuse parts of the input series, while DiffTSST, being generative, sacrifices some CP but excels in SI (Table~\ref{tab:style_integration}), where it consistently captures style characteristics more effectively than baselines. For RM (Table~\ref{tab:realism}), DiffTSST matches or surpasses competing methods, producing outputs that remain close to the real data manifold. In some datasets (e.g., London Smart Meters, Temperature, Wind Farms), Stitching or Wavelets achieve lower overall scores, reflecting their advantage when direct manipulation suffices. However, these methods lack diversity. In contrast, DiffTSST reliably balances content, style, and plausibility while generating novel sequences, underscoring the benefits of diffusion-based style transfer for time series. We conducted these experiments with equal guidance to content and style. We report architectural ablations in Appendix~\ref{app:ablation-arch}, and further experiments in Appendix~\ref{app:ablations_csg} regarding content and style guidance during inference.

\subsection{Downstream Performance: Anomaly Detection in Chemical Data}
We evaluate DiffTSST as a data augmentation method in a low-data anomaly detection setting, using batch distillation as the application domain. Reliable anomaly detection in distillation is essential for maintaining product quality, ensuring safety, and avoiding costly downtime. The dataset is collected from an internal distillation plant contains 200 normal training samples and 100 test samples of key process variables. To augment the training set, we generate 200 additional samples with DiffTSST: each new sequence is synthesized by randomly sampling two normal series from the training set, using one as the content reference and the other as the style reference. The resulting signal combines global dynamics from the content and local variations from the style, thereby enriching the dataset with realistic yet diverse samples.

Anomaly detection is performed using Isolation Forest \citep{liu2008isolation} with a contamination parameter of 0.1. We compare three settings: (i) no augmentation, (ii) augmentation with DiffTSST, and (iii) a control augmentation that creates 200 additional samples by adding small Gaussian noise to existing series. Results in Table~\ref{tab:application} show that DiffTSST-based augmentation yields the best performance (F1 = 0.80, PR-AUC = 0.86), compared to no augmentation (F1 = 0.69, PR-AUC = 0.75) and the noise-based control (F1 = 0.62, PR-AUC = 0.68). These results highlight the importance of structured style transfer for generating meaningful variability, in contrast to naive perturbations that can degrade performance. Further discussion is provided in Appendix~\ref{app:application}.

\begin{table*}[t!]
    \centering
    \caption{Anomaly detection in batch distillation. DiffTSST-based augmentation improves anomaly detection performance compared to no augmentation and noise-based control augmentation.}
    \label{tab:application}
    \begin{tabular}{l|cc|cc|cc}
        \toprule
        \multirow{2}{*}{\textbf{Chemical Process}} & \multicolumn{2}{c|}{\textbf{w/o Aug.}} & \multicolumn{2}{c}{\textbf{With DiffTSST Aug.}} & \multicolumn{2}{c}{\textbf{Control Aug.}} \\
        & F1 $\uparrow$ & PR-AUC $\uparrow$ & F1 $\uparrow$ & PR-AUC $\uparrow$ & F1 $\uparrow$ & PR-AUC $\uparrow$\\
        \midrule
        Batch Distillation & 0.69 & 0.75 & \textbf{0.80} & \textbf{0.86} & 0.62 & 0.68\\
        \bottomrule
    \end{tabular}
\end{table*}

\section{Conclusion}
\label{sec:conclusion}

In this paper, we introduced DiffTSST, a diffusion-based framework for time series style transfer that balances content preservation, style adaptation, and realism. By decomposing time series into content and style components and integrating them within a transformer model, our method enables controlled and flexible style transfer. DiffTSST outperforms baselines across multiple datasets, generalizes to unseen styles via zero-shot transfer, and demonstrates robustness in handling complex temporal patterns. Its effectiveness is further validated in anomaly detection for chemical processes, where it enhances model performance through realistic data augmentation. Beyond style transfer, our approach highlights the potential of diffusion models for structured time series generation. For future work, we plan to enhance DiffTSST with adaptive style interpolation for finer content-style control, domain-specific constraints for better applicability, and few-shot learning to improve adaptability in low-resource scenarios.

\subsubsection*{Acknowledgments}
The main part of this work was conducted within the DFG Research Unit FOR 5359 (ID 459419731) on Deep Learning on Sparse Chemical Process Data. SF and MK further acknowledge support by the DFG through TRR 375 (ID 511263698), SPP 2298 (Id 441826958), and SPP 2331 (441958259), by the Carl-Zeiss Foundation through the initiatives AI-Care and Process Engineering 4.0, and by the BMFTR award 01IS24071A.

\bibliography{references} 
\bibliographystyle{tmlr}

\appendix
\section{Model Architecture}
\label{app:arch}

This appendix specifies the implementation used for all results. The model follows Section~\ref{sec:methodology} and is implemented in a single PyTorch module \texttt{DiffusionTransformer}.

\paragraph{I/O conventions.}
All series are univariate tensors of shape $(B,1,L)$. Training uses fixed windows of $L=128$. When $L$ is not divisible by the patch size $p$, we right-pad to a multiple of $p$ and crop the output back to length $L$.

\subsection*{Content encoder: low-pass, alignment-preserving}
The content encoder $\mathcal{E}_\phi$ removes high-frequency detail while preserving temporal alignment:
\begin{itemize}
\item \textbf{Downsample:} 1D convolution with stride $\mathrm{ds}=8$, kernel $k=5$, GELU.
\item \textbf{Low-res trunk:} $\mathrm{blocks}=3$ conv blocks, each \texttt{Conv1d(H,H,k=5)} $\to$ GELU $\to$ \texttt{Conv1d(H,H,k=5)} $\to$ GELU, with $H=128$ channels.
\item \textbf{Project \& upsample:} $1\times1$ \texttt{Conv1d} to a single channel, then linear interpolation back to the padded input length, followed by cropping to $L$.
\end{itemize}
The output $x^c\in\mathbb{R}^{B\times1\times L}$ is a smoothed, aligned representation.

\subsection*{Style encoder: local, linear-phase, zero-DC}
The style encoder $\mathcal{E}_\psi$ captures local high-frequency structure using small, linear-phase filters that cannot reintroduce offsets or phase shifts. Each \texttt{SymDCFreeConv1d} layer enforces (i) zero DC by subtracting the kernel mean and (ii) symmetry via averaging with the time-reversed kernel; reflection padding avoids edge artifacts. We use:
\[
\texttt{[SymDCFreeConv1d(1,16,k=3) $\to$ GELU] $\times$ 2 $\to$ 1$\times$1 Conv1d(16,1)}.
\]
The output $x^s\in\mathbb{R}^{B\times1\times L}$ emphasizes local fluctuations.

\subsection*{Patchification and unpatching}
We use non-overlapping patches of length $p=8$.
\begin{itemize}
\item \textbf{PatchEmbed:} linear projection from each flattened patch to $d$-dim tokens ($d=256$), producing three token streams: noisy input $E_t$, content $E^c$, style $E^s$.
\item \textbf{Unpatch:} linear projection back to patch space, reshape to $(B,1,L)$, crop to original length.
\end{itemize}

\subsection*{Time conditioning}
The diffusion step $t$ is encoded with a $d$-dimensional sinusoidal embedding followed by a two-layer MLP with SiLU. The resulting vector is added to every token in $E_t$.

\subsection*{Denoiser: stacked self/cross-attention with ALiBi}
The backbone comprises $\mathrm{num\_layers}=4$ identical \emph{DenoisingBlocks}. Each block applies:
\begin{enumerate}
\item self-attention over $E_t$ with ALiBi (additive linear positional biases);
\item cross-attention to content tokens $E^c$ (with ALiBi cross-bias);
\item cross-attention to style tokens $E^s$ (with ALiBi cross-bias);
\item position-wise MLP.
\end{enumerate}
All sublayers use pre-LayerNorm and residual connections. We use $\mathrm{num\_heads}=4$ and hidden size $d=256$.

\subsection*{Classifier-free guidance via input dropout}
During training, we randomly drop the \emph{raw} conditioning streams: with probabilities $p_c=0.10$ (content) and $p_s=0.15$ (style), the respective input is replaced by zeros. The corresponding encoded features are also zeroed to prevent leakage. At inference, we obtain unconditional, content-only, and style-only predictions in a single forward pass and combine them with guidance scales $s_c$ and $s_s$ (Algorithm~\ref{alg:inference}).

\paragraph{Default configuration.}
Unless stated otherwise: $d=256$, $p=8$, $\mathrm{num\_layers}=4$, $\mathrm{num\_heads}=4$, content encoder $(\mathrm{ds}=8, H=128, \mathrm{blocks}=3, k=5)$, style encoder $(\mathrm{hidden}=16, \mathrm{depth}=2, k=3)$. With patch size $p$, the number of tokens is $N=\lceil \frac{L}{p}\rceil$ after padding; ALiBi bias tensors are constructed from $N$ (or $N_c,N_s$) and broadcast across heads and batch.

\section{Training Procedure}
\label{app:train}

\paragraph{Datasets and windowing.}
We train on univariate series from Monash TSF via \texttt{datasets} (the subset names are listed in \texttt{run.py}). Cached raw targets are used; multivariate targets are split into univariate series. Sequences shorter than 128 are discarded. A batch is formed by sampling a series and a random contiguous window of length $L=128$, and z-normalizing the window: $x'=(x-\mu)/(\sigma+10^{-8})$. Missing values are replaced with zeros prior to normalization using \texttt{np.nan\_to\_num}.

\paragraph{Self-supervised conditioning.}
Each iteration uses the same normalized window as source for noisy input, content, and style ($x_0=x_c=x_s$), enabling self-supervised training without paired data.

\paragraph{Forward diffusion.}
We use $T=500$ steps with linear schedule $\beta_t\in[10^{-4},\,0.02]$, $\bar\alpha_t=\prod_{s\le t}(1-\beta_s)$. For $t\sim\{0,\dots,T-1\}$ and $\epsilon\sim\mathcal{N}(0,I)$,
\[
x_t=\sqrt{\bar\alpha_t}\,x_0+\sqrt{1-\bar\alpha_t}\,\epsilon.
\]

\paragraph{Objective and optimization.}
The model predicts $\hat\epsilon=\epsilon_\theta(x_t,t,x_c,x_s)$ and is trained with MSE: $\mathcal{L}=\|\epsilon-\hat\epsilon\|_2^2$. We use AdamW (lr $3\cdot10^{-4}$). Unless noted, batch size is 256 and training runs for 200{,}000 iterations. Classifier-free conditioning uses the input-drop probabilities above.

\paragraph{Logging and checkpoints.}
Every 10 steps we append the loss to a CSV log; every \texttt{log\_interval} steps we save a checkpoint containing iteration number, model state, and optimizer state. Training resumes from the latest checkpoint if present.

\subsection*{Inference}
We generate with DDPM sampling for $T$ steps. At each step we compute $\epsilon_u$ (uncond.), $\epsilon_c$ (content-only), and $\epsilon_s$ (style-only) in a single forward pass by stacking inputs, and combine them
\[
\epsilon=\epsilon_u+s_c(\epsilon_c-\epsilon_u)+s_s(\epsilon_s-\epsilon_u),
\]
then apply the standard DDPM update with posterior variance. A temperature parameter $\lambda$ scales the injected Gaussian noise. Content and style windows are normalized independently; the final output is denormalized with the content statistics.

\section{Hyperparameters}
\label{app:hyper}

Table~\ref{tab:hyperparameters-new} summarizes the default hyperparameters used for DiffTSST. The architectural choices are deliberately lightweight, with a hidden size of 256, four attention heads, and four denoising blocks, striking a balance between expressiveness and computational efficiency. The encoders are specialized: the content encoder applies strided convolutions to capture coarse temporal structure, while the style encoder employs symmetry-preserving convolutions to extract finer-scale variations. Both are coupled through ALiBi positional biases, enabling efficient handling of varying sequence lengths.

For the diffusion process, we adopt 500 steps with a linear $\beta$ schedule, which provides a good trade-off between training stability and sampling speed. Training is performed with AdamW and a batch size of 256 over 200k iterations, with z-score normalization applied per window to stabilize optimization. NaNs are replaced with zeros prior to normalization, ensuring robust handling of real-world time series data.

During inference, guidance scales for both content and style are typically set to one, with the temperature controlling stochasticity in generation. By default, we use $\lambda=1.0$, but as shown in Fig.~\ref{fig:pca-difftsst-temperature}, adjusting this parameter directly modulates diversity in the generated outputs. Together, these hyperparameters form a practical and reproducible default configuration, balancing fidelity, diversity, and efficiency across datasets.

\begin{table}[h]
\centering
\renewcommand{\arraystretch}{1.15}
\begin{tabular}{ll}
\toprule
\multicolumn{2}{c}{\textbf{Architecture}}\\
\midrule
Hidden size ($d$) & 256\\
Attention heads & 4\\
Denoising blocks & 4\\
Patch size ($p$) & 8\\
Time embedding & sinusoidal($d$) $\to$ MLP (SiLU)\\
Content encoder & stride-8 Conv1d ($k=5$), 3 conv blocks, $1\times1$ Conv1d\\
Style encoder & 2 $\times$ [SymDCFreeConv1d ($k=3$), GELU] $\to$ $1\times1$ Conv1d\\
Positional bias & ALiBi (self and cross)\\
CFG drop probs & $p_c=0.10$ (content), $p_s=0.15$ (style)\\
\midrule
\multicolumn{2}{c}{\textbf{Diffusion}}\\
\midrule
Steps $T$ & 500\\
$\beta$ schedule & linear from $10^{-4}$ to $2\cdot10^{-2}$\\
Objective & MSE on added noise\\
\midrule
\multicolumn{2}{c}{\textbf{Training}}\\
\midrule
Iterations & 200{,}000\\
Batch size & 256\\
Optimizer & AdamW (lr $3\cdot10^{-4}$)\\
Window length $L$ & 128\\
Normalization & per-window z-score; NaNs $\to$ 0 before norm\\
Datasets & Monash TSF subsets (listed in \texttt{run.py})\\
Checkpointing & save every \texttt{log\_interval} steps; CSV loss log\\
\midrule
\multicolumn{2}{c}{\textbf{Inference}}\\
\midrule
Guidance scales & $s_c=1.0$, $s_s=1.0$ (unless noted)\\
Temperature & $\lambda=1.0$\\
Length constraint & $L$ multiple of patch size ($p=8$)\\
\bottomrule
\end{tabular}
\caption{Default hyperparameters for DiffTSST.}
\label{tab:hyperparameters-new}
\end{table}

\section{Evaluation Metrics}
\label{app:eval}

We evaluate style transfer with \emph{content preservation} (CP), \emph{style integration} (SI), and \emph{realism} (RM). All metrics are computed on z-normalized windows of equal length.

\subsection*{Multi-scale content/style decomposition}
Given $x\in\mathbb{R}^L$, apply averaging filters with kernels $K=(3,5,15)$ to obtain a smoothed content signal $C(x)$ and residual style signal $S(x)$. Let $x^{(0)}_{\text{smooth}}:=x$ and, for $i=1,2,3$,
\begin{align*}
x^{(i)}_{\text{smooth}} &= \operatorname{AvgFilter}_{K_i}\!\bigl(x^{(i-1)}_{\text{smooth}}\bigr),\\
x^{(i)}_{\text{res}} &= x^{(i-1)}_{\text{smooth}} - x^{(i)}_{\text{smooth}}.
\end{align*}
Set $C(x):=x^{(3)}_{\text{smooth}}$ and $S(x):=\sum_{i=1}^3 x^{(i)}_{\text{res}}$.

\subsection*{Metrics}
Let $\hat{x}$ be the generated series for content $a$ and style $b$ (all of length $L$):
\begin{align*}
\text{CP}(\hat{x},a) &= \tfrac{1}{L}\sum_{t=1}^L\bigl(C(\hat{x})_t - C(a)_t\bigr)^2,\\
\text{SI}(\hat{x},b) &= \tfrac{1}{L}\sum_{t=1}^L\bigl(S(\hat{x})_t - S(b)_t\bigr)^2.
\end{align*}
For realism we embed content and style components with a time-series foundation model $f(\cdot)$ and compute
\[
\text{RM}(\hat{x},a,b)=\tfrac{1}{2}\Bigl[\operatorname{MSE}\bigl(f(C(\hat{x})),f(C(a))\bigr)+\operatorname{MSE}\bigl(f(S(\hat{x})),f(S(b))\bigr)\Bigr].
\]
We report per-dataset means and standard errors across test pairs, and the overall mean of CP, SI, and RM per sample (lower is better for all).

\section{Ablations}
\label{app:ablations}
In this section we conduct ablations regarding DiffTSST architecture, diversity of the generated output, and the guidance parameters of content and style.

\subsection{Diversity}
\label{app:ablations_diversity}
\begin{figure}[t]
  \centering
  \includegraphics[width=0.72\linewidth]{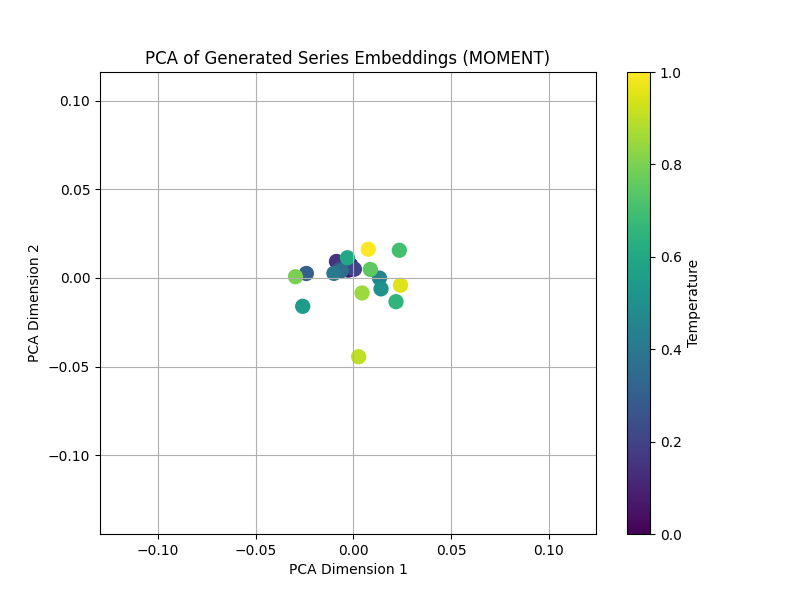}
  \caption{PCA of embeddings of style-transferred series at different temperatures. Higher temperatures lead to greater dispersion along the PCA axes, indicating increased diversity in the generated outputs while remaining within the manifold of realistic series.}
  \label{fig:pca-difftsst-temperature}
\end{figure}

To analyze the effect of sampling temperature, we generate 20 style-transferred series from the same content–style pair, sweeping the temperature from 0 to 1. The resulting embeddings, projected onto the first two PCA components, are shown in Fig.~\ref{fig:pca-difftsst-temperature}. At low temperatures, the samples cluster tightly, indicating conservative modifications that closely preserve the original dynamics. As the temperature increases, the points spread further along the PCA axes, reflecting stronger perturbations and greater diversity in the generated outputs. This trend confirms that \textsc{DiffTSST} can balance realism and variability: while higher temperatures introduce more stylistic variation, the generated series remain coherent and consistent with the embedding space of real data.

\subsection{Architectural Ablations}
\label{app:ablation-arch}

To isolate the contribution of the specialized content and style encoders in DiffTSST, 
we replace them with generic convolutional layers of kernel size $3$. 
This preserves model capacity but removes the carefully designed inductive biases: 
the content encoder’s low-pass filtering for global structure, 
and the style encoder’s zero-DC, symmetry-preserving convolutions for high-frequency variability. 
The results are reported in Table~\ref{tab:ablation-arch}.  

\paragraph{Full DiffTSST.}  
The complete model with both specialized encoders achieves the best overall balance across content preservation (CP), style integration (SI), and realism (RM). This confirms that explicitly disentangling content and style through tailored encoders 
provides the right inductive bias for the task.  

\paragraph{Style encoder $\rightarrow$ Conv1d.}  
When the style encoder is replaced with a plain Conv1d layer, the model can no longer isolate high-frequency variability. As a result, SI worsens considerably, while CP appears to improve because the model primarily relies on the content stream. 
This shows that disentangling style information requires more than generic local filters.  

\paragraph{Content \& style encoders $\rightarrow$ Conv1d.}  When both encoders are replaced with Conv1d layers, the model loses nearly all disentanglement ability: CP, SI, and RM all deteriorate. Without inductive biases, information from content and style streams is blended during training, leading to leakage and poor controllability.  

\paragraph{Conclusion.}  
These ablations demonstrate that the inductive biases encoded in the specialized content and style encoders are crucial. They not only improve SI without sacrificing CP but also stabilize realism, leading to effective and controllable time-series style transfer.

\begin{table}[t]
\centering
\small
\begin{tabular}{l|cccc}
\toprule
Architecture Variant & CP $\downarrow$ & SI $\downarrow$ & RM $\downarrow$ & Avg $\downarrow$ \\
\midrule
DiffTSST (full, specialized encoders) & 0.06 & 0.19 & 0.04 & 0.10 \\
Style Encoder $\rightarrow$ Conv1d($k{=}3$) & 0.02 & 0.36 & 0.06 & 0.15 \\
Content \& Style Encoders $\rightarrow$ Conv1d($k{=}3$) & 0.26 & 0.65 & 0.12 & 0.34 \\
\bottomrule
\end{tabular}
\caption{Architectural ablations of DiffTSST with convolutional replacements. 
To test the effect of specialized encoders, we replace them with plain Conv1d layers (kernel size 3), thereby removing the inductive biases for disentangling content and style while keeping model capacity comparable. 
Replacing the style encoder degrades style integration (SI), as the model defaults to extracting variability from the content stream, while content preservation (CP) remains strong. 
Replacing both encoders with Conv1d layers further harms all metrics, especially SI and realism (RM), confirming that inductive biases in the specialized encoders are essential for effective disentanglement and high-quality style transfer.}
\label{tab:ablation-arch}
\end{table}

\subsection{Content and Style Guidance}
\label{app:ablations_csg}
We study how classifier-free guidance scales affect the trade-off between content preservation (CP), style integration (SI), and realism (RM). Table~\ref{tab:ablation-guidance} varies the content guidance \(s_c\) and style guidance \(s_s\).

\paragraph{Observations.}
(1) \emph{Trade-off axis.} Skewed settings emphasize one objective at the expense of the other: high \(s_c\), low \(s_s\) (row \(1.00/0.25\)) yields strong content adherence (low CP) but weak style integration (high SI), while low \(s_c\), high \(s_s\) (row \(0.25/1.00\)) flips this behavior.
(2) \emph{Balanced sweet spot.} Equal, moderate guidance performs best overall: \(s_c=s_s=1.0\) minimizes the average score. Slightly stronger guidance (\(1.25/1.25\)) is competitive, but pushing both higher degrades CP and/or SI.
(3) \emph{Practical tuning.} For content-critical use, choose \(s_c \gtrsim 1\), \(s_s \lesssim 1\); for style-critical augmentation, invert the ratio. Use the sampling temperature \(\lambda\) to modulate diversity \emph{orthogonally} to guidance (cf.\ Fig.~\ref{fig:pca-difftsst-temperature}): increasing \(\lambda\) spreads samples in PCA space without necessarily changing the CP/SI balance.

\begin{table}[t]
\centering
\small
\begin{tabular}{cc|cccc}
\toprule
$s_c$ & $s_s$ & CP $\downarrow$ & SI $\downarrow$ & RM $\downarrow$ & Avg $\downarrow$ \\
\midrule
1.00 & 0.25 & 0.0653 & 1.0933 & 0.1133 & 0.4240 \\
0.25 & 1.00 & 0.4906 & 0.0469 & 0.0339 & 0.1905 \\
0.25 & 0.25 & 0.4299 & 0.9164 & 0.0994 & 0.4819 \\
0.50 & 0.50 & 0.1065 & 0.6837 & 0.0878 & 0.2927 \\
0.75 & 0.75 & 0.0643 & 0.4786 & 0.0587 & 0.2006 \\
\textbf{1.00} & \textbf{1.00} & \textbf{0.1283} & \textbf{0.0780} & \textbf{0.0191} & \textbf{0.0751} \\
1.25 & 1.25 & 0.2207 & 0.0772 & 0.0184 & 0.1054 \\
1.50 & 1.50 & 0.2465 & 0.1519 & 0.0281 & 0.1421 \\
1.75 & 1.75 & 0.2862 & 0.2753 & 0.0313 & 0.1976 \\
2.00 & 2.00 & 0.2596 & 0.6984 & 0.0392 & 0.3324 \\
\bottomrule
\end{tabular}
\caption{Ablation of classifier-free guidance scales. \(s_c\) steers content adherence; \(s_s\) steers style strength. Lower is better for all metrics; the best Avg occurs at \(s_c=s_s=1.0\).}
\label{tab:ablation-guidance}
\end{table}

\section{Extended Quantitative Evaluation}
\label{app:quantitative}
\noindent\textbf{Reading the tables.} We report means $\pm$ standard errors across test pairs. Lower is better for all metrics. Deterministic baselines (Stitching, Wavelets) often \emph{reuse} parts of the inputs, which favors content preservation but limits diversity; our method is generative and can trade content adherence for better style integration and realism.

\paragraph{Content preservation (Table~\ref{tab:content_preservation}).}
As expected, Stitching, Wavelets, and NST achieve near-zero CP on most datasets because they either directly overlay residuals or optimize to the reference content. DiffTSST shows a small but consistent CP penalty overall ($0.06\pm0.01$) due to being generative and not copying content segments verbatim. The largest CP gaps appear on datasets with strong piecewise-constant structure or long plateaus (\emph{London Smart Meters} $0.19$, \emph{NN5 Daily} $0.18$, \emph{US Births} $0.26$, \emph{Vehicle Trips} $0.19$, \emph{Temperature (rain)} $0.17$), where strict alignment to low-frequency level shifts is particularly hard without sacrificing style. This is a tunable trade-off: increasing the content guidance $s_c$, decreasing the style guidance $s_s$, or lowering the sampling temperature $\lambda$ (Algorithm~\ref{alg:inference}) improves CP at the cost of style strength and diversity.

\paragraph{Style integration (Table~\ref{tab:style_integration}).}
DiffTSST is best (or tied) on the majority of datasets and by a clear margin overall ($0.19\pm0.04$ vs.\ $0.31$/$0.40$/$0.62$). Gains are pronounced where local volatility patterns matter: \emph{NN5 Daily} ($0.15$ vs.\ $0.50$/$0.70$/$1.28$), \emph{US Births} ($0.16$ vs.\ $0.51$/$0.78$/$1.14$), \emph{Vehicle Trips} ($0.13$ vs.\ $0.48$/$0.70$/$1.07$), \emph{KDD Cup 2018} ($0.20$ vs.\ $0.30$/$0.36$/$0.55$), and \emph{Pedestrian Counts} ($0.23$ vs.\ $0.41$/$0.50$/$0.71$). On very high-frequency signals (\emph{Solar (4sec)}, \emph{Wind (4sec)}) the gap narrows, and methods are effectively tied, indicating that simple residual swapping is sometimes adequate when style is dominated by white-like noise. Nevertheless, DiffTSST remains competitive even on challenging settings with mixed temporal scales (\emph{Traffic (hourly)}: $0.51$, the best among methods though absolute errors remain high for all).

\paragraph{Realism (Table~\ref{tab:realism}).}
All methods achieve similar RM overall (Stitching $0.04$, Wavelets $0.05$, NST $0.05$, DiffTSST $0.04$). Despite not copying content, DiffTSST matches the best baselines on average and is often tied within rounding on individual datasets (e.g., \emph{London Smart Meters}, \emph{Tourism (monthly)}, \emph{Weather}). Stitching sometimes has a slight edge on domains where simple residual recombination stays close to the training manifold (\emph{Rideshare}), but DiffTSST frequently attains the lowest RM as well (\emph{Bitcoin}, \emph{KDD Cup 2018}, \emph{Solar (4sec)}, \emph{Wind (4sec)}). The parity in RM, together with superior SI, indicates that our generations remain statistically plausible while achieving stronger stylistic fidelity.

\paragraph{Takeaways.}
(1) DiffTSST delivers substantially better style integration without sacrificing realism. (2) A modest CP gap is an inherent result of generative sampling rather than content copying and can be reduced by adjusting guidance $(s_c,s_s)$ and temperature $\lambda$. (3) On purely high-frequency styles, simple baselines can approach DiffTSST, but on mixed-scale and structured styles, diffusion with content/style conditioning provides clear advantages.

\begin{table*}[ht]
\centering
\caption{Evaluation (Content Preservation (CP)). Lower is better.}
\label{tab:content_preservation}
\begin{tabular}{l|cccc}
\hline
Dataset & Stitching & Wavelets & NST & DiffTSST \\
\hline
Electricity  Demand & $0.00 \, \pm \, 0.00$ & $0.00 \, \pm \, 0.00$ & $\mathbf{0.00} \, \pm \, \mathbf{0.00}$ & $0.00 \, \pm \, 0.00$ \\
Bitcoin & $0.00 \, \pm \, 0.00$ & $\mathbf{0.00} \, \pm \, \mathbf{0.00}$ & $0.00 \, \pm \, 0.00$ & $0.01 \, \pm \, 0.00$ \\
Covid Deaths & $0.00 \, \pm \, 0.00$ & $0.00 \, \pm \, 0.00$ & $\mathbf{0.00} \, \pm \, \mathbf{0.00}$ & $0.01 \, \pm \, 0.00$ \\
FRED-MD & $0.00 \, \pm \, 0.00$ & $\mathbf{0.00} \, \pm \, \mathbf{0.00}$ & $0.00 \, \pm \, 0.00$ & $0.00 \, \pm \, 0.00$ \\
Kaggle Web Traffic & $0.00 \, \pm \, 0.00$ & $\mathbf{0.00} \, \pm \, \mathbf{0.00}$ & $0.00 \, \pm \, 0.00$ & $0.01 \, \pm \, 0.00$ \\
KDD Cup 2018 & $0.00 \, \pm \, 0.00$ & $\mathbf{0.00} \, \pm \, \mathbf{0.00}$ & $0.00 \, \pm \, 0.00$ & $0.02 \, \pm \, 0.00$ \\
London Smart Meters & $0.00 \, \pm \, 0.00$ & $0.00 \, \pm \, 0.00$ & $\mathbf{0.00} \, \pm \, \mathbf{0.00}$ & $0.19 \, \pm \, 0.00$ \\
NN5 Daily & $0.00 \, \pm \, 0.00$ & $\mathbf{0.00} \, \pm \, \mathbf{0.00}$ & $0.00 \, \pm \, 0.00$ & $0.18 \, \pm \, 0.00$ \\
Oikolab Weather & $0.01 \, \pm \, 0.00$ & $0.00 \, \pm \, 0.00$ & $\mathbf{0.00} \, \pm \, \mathbf{0.00}$ & $0.06 \, \pm \, 0.00$ \\
Pedestrian Counts & $0.00 \, \pm \, 0.00$ & $\mathbf{0.00} \, \pm \, \mathbf{0.00}$ & $0.00 \, \pm \, 0.00$ & $0.03 \, \pm \, 0.00$ \\
Rideshare & $0.00 \, \pm \, 0.00$ & $\mathbf{0.00} \, \pm \, \mathbf{0.00}$ & $0.00 \, \pm \, 0.00$ & $0.03 \, \pm \, 0.00$ \\
Saugeenday & $0.00 \, \pm \, 0.00$ & $\mathbf{0.00} \, \pm \, \mathbf{0.00}$ & $0.00 \, \pm \, 0.00$ & $0.01 \, \pm \, 0.00$ \\
Solar (10min) & $0.00 \, \pm \, 0.00$ & $0.00 \, \pm \, 0.00$ & $\mathbf{0.00} \, \pm \, \mathbf{0.00}$ & $0.00 \, \pm \, 0.00$ \\
Solar (4sec) & $0.00 \, \pm \, 0.00$ & $\mathbf{0.00} \, \pm \, \mathbf{0.00}$ & $0.00 \, \pm \, 0.00$ & $0.00 \, \pm \, 0.00$ \\
Sunspot & $0.00 \, \pm \, 0.00$ & $\mathbf{0.00} \, \pm \, \mathbf{0.00}$ & $0.00 \, \pm \, 0.00$ & $0.05 \, \pm \, 0.00$ \\
Temperature (rain) & $0.00 \, \pm \, 0.00$ & $\mathbf{0.00} \, \pm \, \mathbf{0.00}$ & $0.00 \, \pm \, 0.00$ & $0.17 \, \pm \, 0.00$ \\
Tourism (monthly) & $0.00 \, \pm \, 0.00$ & $0.00 \, \pm \, 0.00$ & $\mathbf{0.00} \, \pm \, \mathbf{0.00}$ & $0.09 \, \pm \, 0.00$ \\
Traffic (hourly) & $0.00 \, \pm \, 0.00$ & $0.00 \, \pm \, 0.00$ & $\mathbf{0.00} \, \pm \, \mathbf{0.00}$ & $0.04 \, \pm \, 0.00$ \\
US Births & $0.00 \, \pm \, 0.00$ & $0.00 \, \pm \, 0.00$ & $\mathbf{0.00} \, \pm \, \mathbf{0.00}$ & $0.26 \, \pm \, 0.00$ \\
Vehicle Trips & $0.00 \, \pm \, 0.00$ & $0.00 \, \pm \, 0.00$ & $\mathbf{0.00} \, \pm \, \mathbf{0.00}$ & $0.19 \, \pm \, 0.00$ \\
Weather & $0.00 \, \pm \, 0.00$ & $0.00 \, \pm \, 0.00$ & $\mathbf{0.00} \, \pm \, \mathbf{0.00}$ & $0.08 \, \pm \, 0.00$ \\
Wind (4sec) & $0.00 \, \pm \, 0.00$ & $0.00 \, \pm \, 0.00$ & $\mathbf{0.00} \, \pm \, \mathbf{0.00}$ & $0.00 \, \pm \, 0.00$ \\
Wind Farms (min) & $0.00 \, \pm \, 0.00$ & $\mathbf{0.00} \, \pm \, \mathbf{0.00}$ & $0.00 \, \pm \, 0.00$ & $0.05 \, \pm \, 0.01$ \\
\hline
Overall & $0.00 \, \pm \, 0.00$ & $0.00 \, \pm \, 0.00$ & $0.00 \, \pm \, 0.00$ & $0.06 \, \pm \, 0.01$ \\
\hline
\end{tabular}
\end{table*}

\begin{table*}[ht]
\centering
\caption{Evaluation (Style Integration (SI)). Lower is better.}
\label{tab:style_integration}
\begin{tabular}{l|cccc}
\hline
Dataset & Stitching & Wavelets & NST & DiffTSST \\
\hline
Electricity  Demand & $0.18 \, \pm \, 0.06$ & $0.20 \, \pm \, 0.07$ & $0.32 \, \pm \, 0.08$ & $\mathbf{0.18} \, \pm \, \mathbf{0.06}$ \\
Bitcoin & $0.14 \, \pm \, 0.01$ & $0.16 \, \pm \, 0.02$ & $0.27 \, \pm \, 0.02$ & $\mathbf{0.11} \, \pm \, \mathbf{0.02}$ \\
Covid Deaths & $0.19 \, \pm \, 0.04$ & $0.23 \, \pm \, 0.06$ & $0.34 \, \pm \, 0.06$ & $\mathbf{0.16} \, \pm \, \mathbf{0.03}$ \\
FRED-MD & $0.08 \, \pm \, 0.01$ & $0.09 \, \pm \, 0.01$ & $0.20 \, \pm \, 0.01$ & $\mathbf{0.08} \, \pm \, \mathbf{0.01}$ \\
Kaggle Web Traffic & $0.19 \, \pm \, 0.01$ & $0.22 \, \pm \, 0.02$ & $0.34 \, \pm \, 0.02$ & $\mathbf{0.12} \, \pm \, \mathbf{0.01}$ \\
KDD Cup 2018 & $0.30 \, \pm \, 0.03$ & $0.36 \, \pm \, 0.03$ & $0.55 \, \pm \, 0.05$ & $\mathbf{0.20} \, \pm \, \mathbf{0.01}$ \\
London Smart Meters & $0.42 \, \pm \, 0.01$ & $0.48 \, \pm \, 0.01$ & $0.83 \, \pm \, 0.01$ & $\mathbf{0.33} \, \pm \, \mathbf{0.03}$ \\
NN5 Daily & $0.50 \, \pm \, 0.06$ & $0.70 \, \pm \, 0.09$ & $1.28 \, \pm \, 0.09$ & $\mathbf{0.15} \, \pm \, \mathbf{0.02}$ \\
Oikolab Weather & $0.39 \, \pm \, 0.05$ & $0.52 \, \pm \, 0.07$ & $0.67 \, \pm \, 0.13$ & $\mathbf{0.26} \, \pm \, \mathbf{0.04}$ \\
Pedestrian Counts & $0.41 \, \pm \, 0.01$ & $0.50 \, \pm \, 0.01$ & $0.71 \, \pm \, 0.01$ & $\mathbf{0.23} \, \pm \, \mathbf{0.01}$ \\
Rideshare & $0.42 \, \pm \, 0.10$ & $0.51 \, \pm \, 0.12$ & $0.71 \, \pm \, 0.15$ & $\mathbf{0.34} \, \pm \, \mathbf{0.05}$ \\
Saugeenday & $0.23 \, \pm \, 0.02$ & $0.26 \, \pm \, 0.02$ & $0.39 \, \pm \, 0.02$ & $\mathbf{0.14} \, \pm \, \mathbf{0.01}$ \\
Solar (10min) & $0.17 \, \pm \, 0.04$ & $0.20 \, \pm \, 0.04$ & $0.30 \, \pm \, 0.05$ & $\mathbf{0.16} \, \pm \, \mathbf{0.05}$ \\
Solar (4sec) & $\mathbf{0.10} \, \pm \, \mathbf{0.02}$ & $0.11 \, \pm \, 0.03$ & $0.21 \, \pm \, 0.03$ & $0.10 \, \pm \, 0.03$ \\
Sunspot & $0.52 \, \pm \, 0.08$ & $0.62 \, \pm \, 0.12$ & $0.87 \, \pm \, 0.18$ & $\mathbf{0.29} \, \pm \, \mathbf{0.05}$ \\
Temperature (rain) & $0.47 \, \pm \, 0.02$ & $0.74 \, \pm \, 0.02$ & $1.18 \, \pm \, 0.02$ & $\mathbf{0.30} \, \pm \, \mathbf{0.04}$ \\
Tourism (monthly) & $0.30 \, \pm \, 0.01$ & $0.38 \, \pm \, 0.01$ & $0.83 \, \pm \, 0.01$ & $\mathbf{0.14} \, \pm \, \mathbf{0.04}$ \\
Traffic (hourly) & $0.69 \, \pm \, 0.09$ & $0.80 \, \pm \, 0.10$ & $0.93 \, \pm \, 0.11$ & $\mathbf{0.51} \, \pm \, \mathbf{0.07}$ \\
US Births & $0.51 \, \pm \, 0.03$ & $0.78 \, \pm \, 0.03$ & $1.14 \, \pm \, 0.03$ & $\mathbf{0.16} \, \pm \, \mathbf{0.05}$ \\
Vehicle Trips & $0.48 \, \pm \, 0.03$ & $0.70 \, \pm \, 0.04$ & $1.07 \, \pm \, 0.04$ & $\mathbf{0.13} \, \pm \, \mathbf{0.03}$ \\
Weather & $0.31 \, \pm \, 0.01$ & $0.43 \, \pm \, 0.01$ & $0.66 \, \pm \, 0.01$ & $\mathbf{0.18} \, \pm \, \mathbf{0.03}$ \\
Wind (4sec) & $0.12 \, \pm \, 0.03$ & $0.13 \, \pm \, 0.04$ & $0.23 \, \pm \, 0.04$ & $\mathbf{0.12} \, \pm \, \mathbf{0.03}$ \\
Wind Farms (min) & $0.08 \, \pm \, 0.01$ & $0.09 \, \pm \, 0.01$ & $0.16 \, \pm \, 0.01$ & $\mathbf{0.08} \, \pm \, \mathbf{0.01}$ \\
\hline
Overall & $0.31 \, \pm \, 0.06$ & $0.40 \, \pm \, 0.10$ & $0.62 \, \pm \, 0.17$ & $0.19 \, \pm \, 0.04$ \\
\hline
\end{tabular}
\end{table*}

\begin{table*}[ht]
\centering
\caption{Evaluation (Realism (RM)). Lower is better.}
\label{tab:realism}
\begin{tabular}{l|cccc}
\hline
Dataset & Stitching & Wavelets & NST & DiffTSST \\
\hline
Electricity  Demand & $\mathbf{0.03} \, \pm \, \mathbf{0.00}$ & $0.04 \, \pm \, 0.00$ & $0.05 \, \pm \, 0.00$ & $0.03 \, \pm \, 0.00$ \\
Bitcoin & $0.04 \, \pm \, 0.00$ & $0.04 \, \pm \, 0.00$ & $0.06 \, \pm \, 0.00$ & $\mathbf{0.03} \, \pm \, \mathbf{0.00}$ \\
Covid Deaths & $0.03 \, \pm \, 0.00$ & $0.05 \, \pm \, 0.00$ & $0.06 \, \pm \, 0.00$ & $\mathbf{0.03} \, \pm \, \mathbf{0.00}$ \\
FRED-MD & $0.04 \, \pm \, 0.00$ & $0.05 \, \pm \, 0.00$ & $0.06 \, \pm \, 0.00$ & $\mathbf{0.03} \, \pm \, \mathbf{0.00}$ \\
Kaggle Web Traffic & $0.04 \, \pm \, 0.00$ & $0.04 \, \pm \, 0.00$ & $0.05 \, \pm \, 0.00$ & $\mathbf{0.03} \, \pm \, \mathbf{0.00}$ \\
KDD Cup 2018 & $0.03 \, \pm \, 0.00$ & $0.03 \, \pm \, 0.00$ & $0.04 \, \pm \, 0.00$ & $\mathbf{0.03} \, \pm \, \mathbf{0.00}$ \\
London Smart Meters & $0.05 \, \pm \, 0.00$ & $0.04 \, \pm \, 0.00$ & $\mathbf{0.04} \, \pm \, \mathbf{0.00}$ & $0.04 \, \pm \, 0.00$ \\
NN5 Daily & $0.06 \, \pm \, 0.00$ & $0.07 \, \pm \, 0.00$ & $0.06 \, \pm \, 0.00$ & $\mathbf{0.04} \, \pm \, \mathbf{0.00}$ \\
Oikolab Weather & $0.03 \, \pm \, 0.00$ & $0.04 \, \pm \, 0.00$ & $0.04 \, \pm \, 0.00$ & $\mathbf{0.03} \, \pm \, \mathbf{0.00}$ \\
Pedestrian Counts & $0.06 \, \pm \, 0.00$ & $0.05 \, \pm \, 0.00$ & $0.05 \, \pm \, 0.00$ & $\mathbf{0.05} \, \pm \, \mathbf{0.00}$ \\
Rideshare & $\mathbf{0.04} \, \pm \, \mathbf{0.00}$ & $0.05 \, \pm \, 0.00$ & $0.05 \, \pm \, 0.00$ & $0.05 \, \pm \, 0.00$ \\
Saugeenday & $\mathbf{0.03} \, \pm \, \mathbf{0.00}$ & $0.03 \, \pm \, 0.00$ & $0.05 \, \pm \, 0.00$ & $0.04 \, \pm \, 0.00$ \\
Solar (10min) & $0.04 \, \pm \, 0.00$ & $0.05 \, \pm \, 0.00$ & $0.06 \, \pm \, 0.00$ & $\mathbf{0.03} \, \pm \, \mathbf{0.00}$ \\
Solar (4sec) & $\mathbf{0.03} \, \pm \, \mathbf{0.00}$ & $0.04 \, \pm \, 0.00$ & $0.07 \, \pm \, 0.00$ & $0.03 \, \pm \, 0.00$ \\
Sunspot & $\mathbf{0.03} \, \pm \, \mathbf{0.00}$ & $0.03 \, \pm \, 0.00$ & $0.03 \, \pm \, 0.00$ & $0.04 \, \pm \, 0.00$ \\
Temperature (rain) & $0.05 \, \pm \, 0.00$ & $\mathbf{0.05} \, \pm \, \mathbf{0.00}$ & $0.07 \, \pm \, 0.00$ & $0.05 \, \pm \, 0.00$ \\
Tourism (monthly) & $0.04 \, \pm \, 0.00$ & $\mathbf{0.04} \, \pm \, \mathbf{0.00}$ & $0.05 \, \pm \, 0.00$ & $0.05 \, \pm \, 0.00$ \\
Traffic (hourly) & $\mathbf{0.04} \, \pm \, \mathbf{0.00}$ & $0.04 \, \pm \, 0.00$ & $0.04 \, \pm \, 0.00$ & $0.05 \, \pm \, 0.00$ \\
US Births & $0.08 \, \pm \, 0.00$ & $0.07 \, \pm \, 0.00$ & $0.07 \, \pm \, 0.00$ & $\mathbf{0.05} \, \pm \, \mathbf{0.00}$ \\
Vehicle Trips & $0.08 \, \pm \, 0.00$ & $0.07 \, \pm \, 0.00$ & $0.07 \, \pm \, 0.00$ & $\mathbf{0.04} \, \pm \, \mathbf{0.00}$ \\
Weather & $0.04 \, \pm \, 0.00$ & $\mathbf{0.04} \, \pm \, \mathbf{0.00}$ & $0.04 \, \pm \, 0.00$ & $0.04 \, \pm \, 0.00$ \\
Wind (4sec) & $\mathbf{0.03} \, \pm \, \mathbf{0.00}$ & $0.04 \, \pm \, 0.00$ & $0.06 \, \pm \, 0.00$ & $0.03 \, \pm \, 0.00$ \\
Wind Farms (min) & $0.07 \, \pm \, 0.00$ & $0.17 \, \pm \, 0.00$ & $0.09 \, \pm \, 0.00$ & $\mathbf{0.07} \, \pm \, \mathbf{0.00}$ \\
\hline
Overall & $0.04 \, \pm \, 0.00$ & $0.05 \, \pm \, 0.00$ & $0.05 \, \pm \, 0.00$ & $0.04 \, \pm \, 0.00$ \\
\hline
\end{tabular}
\end{table*}

\section{Zero-Shot Length Extrapolation with ALiBi}
\label{app:alibi}

\paragraph{Setup.}
We train the denoiser on fixed windows of length $L{=}128$ using Attention with Linear Biases (ALiBi), which adds a linear distance penalty to attention scores and removes the need for learned absolute position embeddings. To test length extrapolation, we keep model weights and all sampling hyperparameters fixed and run inference at $L\in\{128,256,512,1024,2048\}$. Following our standard TSST pipeline, we condition on a smooth low-frequency content and a high-frequency style sequence; both are synthetic controls to isolate the effect of length scaling. Sampling uses $s_{c}{=}s_{s}{=}1.0$ and temperature $0.0$ for all $L$.

\begin{figure*}[t]
    \centering
    \includegraphics[width=\textwidth]{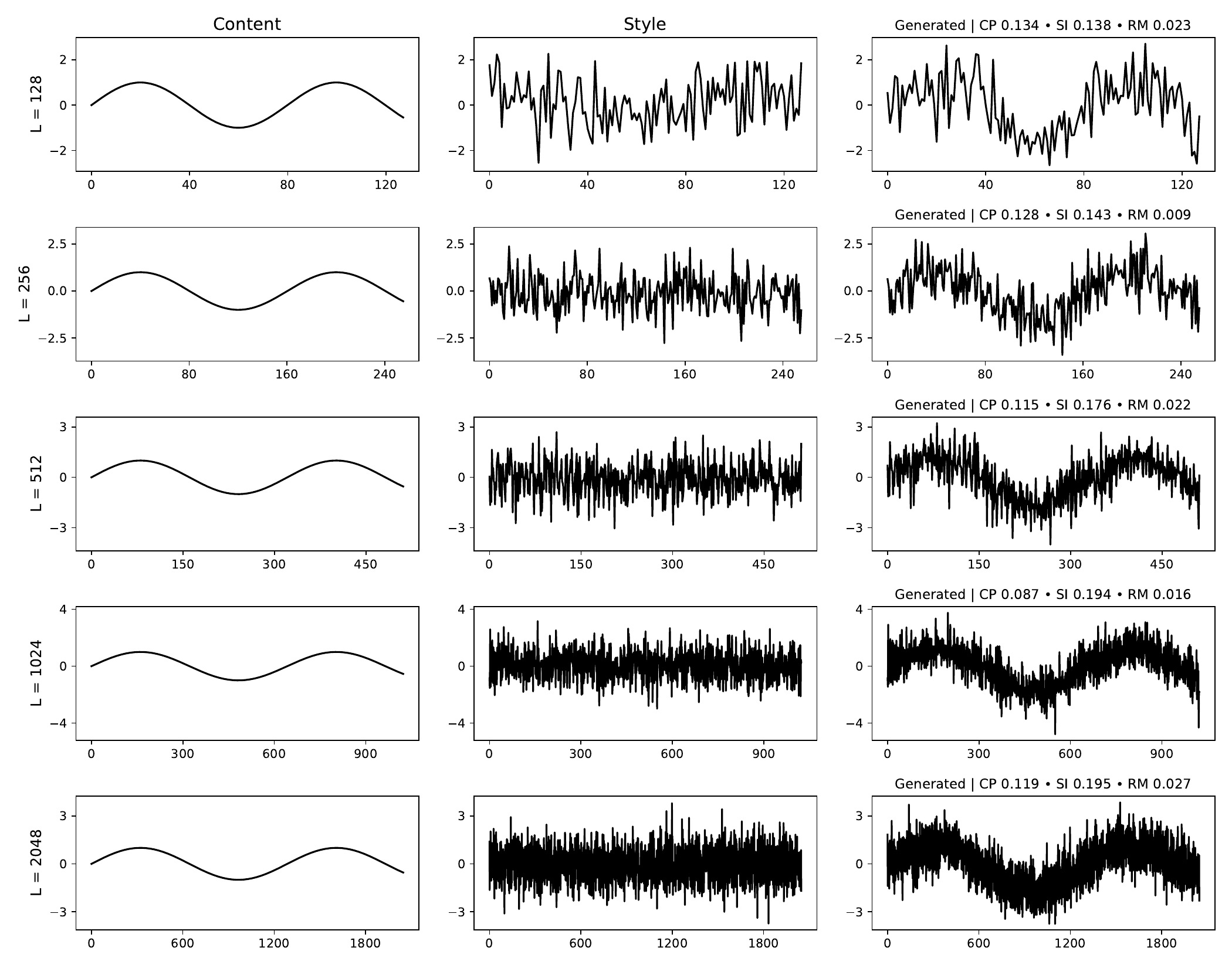}
    \caption{DiffTSST enables zero-shot length extrapolation with stable metrics.
    Rows show increasing sequence lengths $L\in\{128,256,512,1024,2048\}$; columns show the \emph{content} input, the \emph{style} input, and the \emph{generated} output. The generated column annotates CP, SI, and RM for each $L$. The same model, trained only at $L{=}128$, preserves global content shape and consistently integrates high-frequency style across much longer horizons.}
    \label{fig:alibi-long-metrics}
\end{figure*}

\paragraph{Results.}
Table~\ref{tab:alibi-metric-table} lists the quantitative values that appear in Figure~\ref{fig:alibi-long-metrics}. Across $16\times$ longer contexts, all metrics remain in a comparable range to the training length. We observe a gentle trend of slightly \emph{lower} CP (better preservation) up to $L{=}1024$ and a mild rebound at $L{=}2048$, while SI improves gradually with length, suggesting that high-frequency fluctuations are injected uniformly rather than decaying after the training horizon. RM stays low across all $L$, indicating no obvious realism degradation due to extrapolation.

\begin{table}[h]
\centering
\caption{Successful length extrapolation with ALiBi. Style transfer quality remains consistent across longer horizons. Metrics correspond to Fig.~\ref{fig:alibi-long-metrics}.}
\label{tab:alibi-metric-table}
\begin{tabular}{rccc}
\toprule
$L$ & CP $\downarrow$ & SI $\downarrow$ & RM $\downarrow$ \\
\midrule
128  & 0.134 & 0.138 & 0.023 \\
256  & 0.128 & 0.143 & 0.009 \\
512  & 0.115 & 0.176 & 0.022 \\
1024 & 0.087 & 0.194 & 0.016 \\
2048 & 0.119 & 0.195 & 0.027 \\
\bottomrule
\end{tabular}
\end{table}

\section{Application: Anomaly Detection in Chemical Data}
\label{app:application}

As a proof of concept, we evaluate anomaly detection in batch distillation using the Isolation Forest algorithm \citep{liu2008isolation}. Isolation Forest is an unsupervised learning method that detects anomalies by constructing random decision trees and measuring how easily points are isolated. Anomalies, which exhibit unique or sparse patterns, tend to be isolated faster, resulting in a lower average path length. We use a contamination parameter of 0.1, assuming that 10\% of the dataset consists of anomalies.

The experimental data was sourced internally from a batch distillation plant, where key process variables such as temperature, pressure, and concentration profiles were collected over multiple runs. The dataset consists of 200 training samples and 100 test samples. Detecting anomalies in batch distillation is particularly challenging due to process variability, sensor noise, and overlapping distributions between normal and anomalous conditions. Consequently, robust anomaly detection models require diverse and representative training data.

To enhance anomaly detection, we apply DiffTSST as a data augmentation technique, generating 200 additional augmented samples by transferring styles within the experimental dataset. TSST-based augmentation follows a similar approach as in vision and text processing, where generative techniques improve generalization and mitigate biases in training. In computer vision, methods such as style transfer and generative adversarial networks (GANs) synthesize realistic yet diverse samples, while in text processing, data augmentation techniques introduce semantic variability to strengthen model robustness. Inspired by these advancements, we leverage TSST to introduce meaningful variations in time series data while preserving core structural information.

To further analyze the effect of augmentation, we introduce a control augmentation method that generates 200 additional samples by adding small noise to the original data. As shown in Table~\ref{tab:application}, augmentation with DiffTSST significantly improves performance, increasing the F1-score from 0.69 to 0.80 and PR-AUC from 0.75 to 0.86. In contrast, the control augmentation leads to a decline in performance, with an F1-score of 0.62 and PR-AUC of 0.68. The lower performance of the control approach suggests that simply adding small noise fails to provide meaningful diversity, leading to redundant training samples that do not contribute to improved generalization. By creating near-duplicate data points, noise-based augmentation may cause the model to overfit to minor fluctuations rather than learning the broader distribution of normal and anomalous patterns.

DiffTSST, on the other hand, enhances diversity in the training set by generating novel yet realistic variations of existing patterns, helping the model better distinguish anomalies from normal process variations. The method effectively captures process dynamics and introduces controlled variability, reducing overfitting to specific operating conditions and improving adaptability to new, unseen anomalies. These results highlight the advantages of TSST-based augmentation in time series modeling, particularly for industrial applications where data scarcity and process variability present significant challenges.

\end{document}